\pdfoutput=1
\documentclass[11pt]{article}

\usepackage{acl}

\usepackage{times}
\usepackage{latexsym}

\usepackage{tcolorbox}
\tcbuselibrary{breakable}
\usepackage{caption} 
\usepackage{booktabs}
\usepackage{multirow}
\usepackage{xcolor}
\usepackage{colortbl}
\usepackage{array} 
\usepackage{amsmath}
\usepackage{xcolor}
\usepackage{makecell}
\usepackage{subcaption}

\usepackage{tabularx}      
\usepackage{booktabs}      
\usepackage{array}         
\usepackage{enumitem}      
\newcolumntype{Y}{>{\raggedright\arraybackslash}X} 
\usepackage{placeins}

\usepackage{arydshln}
\usepackage{amssymb}
\usepackage{pifont}

\newcolumntype{C}{>{\centering\arraybackslash}p{0.07\textwidth}}
\usepackage[T1]{fontenc}

\usepackage[utf8]{inputenc}

\usepackage{microtype}

\usepackage{inconsolata}

\usepackage{graphicx}

\title{Understanding LLM Reasoning for Abstractive Summarization}

\author{Haohan Yuan \qquad Haopeng Zhang\thanks{corresponding author}\\ 
  ALOHA Lab, University of Hawaii at Manoa \\
  \texttt{\{haohany,haopengz\}@hawaii.edu}}

\begin{document}
\maketitle

\begin{abstract}
Reasoning has substantially improved Large Language Models (LLMs) on analytical tasks such as mathematics and code generation, but its value for abstractive summarization remains unclear. To address this gap, we adapt general reasoning strategies to the summarization setting and conduct a large-scale comparative study of 8 reasoning strategies and 3 Large Reasoning Models (LRMs) across 8 diverse datasets, evaluating both summary quality and factual faithfulness. Our results show that reasoning is not a universal solution and its effectiveness depends strongly on the strategy and the summarization setting. In particular, we find a trade-off between summary quality and factual faithfulness. Explicit reasoning strategies often improve reference-based quality, but may weaken factual grounding, whereas implicit reasoning in LRMs shows the opposite tendency. We further find that increasing an LRM's internal reasoning budget does not reliably improve summarization and can even reduce factual consistency. These findings suggest that, for summarization, more reasoning is not always better. Effective reasoning should preserve faithful compression rather than induce over-elaboration. Our source code is publicly available. \footnote{\url{https://github.com/alohalab-ai/Summarization_reasoning}}
\end{abstract}

\section{Introduction}
LLMs have demonstrated remarkable capabilities across many natural language processing tasks~\cite{zhang-etal-2024-benchmarking,openai2024gpt4technicalreport,yuan2025strucsum,zhang-2025-bridging}. A significant recent development is the paradigm of LLM reasoning, where models are prompted to articulate intermediate thoughts before responding. While techniques such as Chain-of-Thought (CoT)~\cite{wei2022chain} and Self-Consistency~\cite{wang2022self} have established state-of-the-art performance in tasks like mathematics and logical inference~\cite{kojima2022large,huang2023towards,chen2023universal,tan2025scalingbehaviorsllmreinforcement}, their utility for abstractive summarization remains largely assumed rather than proven.

Despite the surge of interest in reasoning, its application to summarization lacks systematic study. Prior work by \citet{adams2023sparse} found that stepwise reasoning can influence the balance between informativeness and readability in news summarization. Similarly, \citet{chang2023booookscore} utilized hierarchical reasoning for book-length texts, while \citet{zhang2023summit} demonstrated the utility of reflective revision and refinement. However, summarization differs fundamentally from logic puzzles, as it requires information compression rather than derivation. Consequently, several critical questions remain unanswered: \textbf{RQ1:} Does reasoning actually improve summarization performance compared to standard prompting?  
\textbf{RQ2:} When do reasoning strategies excel across different datasets and context?
\textbf{RQ3:} How do different reasoning methods impact summarization results?

To bridge this gap, we present the first large-scale, systematic evaluation of reasoning methods in LLM-based summarization. We establish a unified evaluation framework to assess eight representative strategies, categorized into three paradigms:
\textit{Augmentation-based} methods, including Chain-of-Thought (CoT), Cited Summarization (Cite), Extract-to-Abstract (E2A), and Question-Answer Guided (QAG); 
\textit{Organization-based} methods, including Decomposition (Deco) and Plan-then-Write (Plan); 
and \textit{Reflective} methods, including Iterative Refine (IR) and Self-Consistency (SC). We propose task-specific adaptations of Cite, QAG, Plan, and SC for abstractive summarization (via prompts and rubrics) and evaluate them within a unified framework. Furthermore, we benchmark these explicit reasoning methods against three leading Large Reasoning Models (LRMs)~\cite{openai2024openaio1card}: \textsc{o1}, \textsc{o3}, and \textsc{GPT-5}. 

To ensure robustness, our evaluation spans eight datasets covering diverse domains, lengths, and formats, including news, dialogue, social media, knowledge bases, scientific writing, narrative novels, and structured tables.
We employ a comprehensive suite of metrics: reference-based quality (ROUGE~\cite{lin-2004-rouge}, BERTScore~\cite{zhang2019bertscore}), factual consistency (SummaC~\cite{laban-etal-2022-summac}, AlignScore~\cite{zha2023alignscore}), LLM-as-a-judge (G-Eval~\cite{liu2023geval}), and human evaluation.

Our empirical analysis yields counter-intuitive findings that challenge prevailing assumptions. 
First, \textbf{reasoning is not a panacea}. A simple \textit{Vanilla} prompt often matches or surpasses complex reasoning strategies, particularly in few-shot settings. 
Second, we observe a distinct \textbf{trade-off between quality and faithfulness}. Explicit reasoning strategies like SC and IR tend to improve linguistic fluency and reference-based scores, but often at the cost of hallucination. Conversely, implicit reasoning within LRMs excels at factual grounding but may sacrifice stylistic polish. We further note that LLM-as-a-judge metrics tend to overestimate faithfulness compared to human evaluation.

The contributions of this work are fourfold:
\textbf{(1)} We adapt general reasoning paradigms (Cite, QAG, Plan, and SC) to the context of abstractive summarization. 
\textbf{(2)} We establish the first systematic evaluation for reasoning in summarization, encompassing 8 explicit methods, 3 LRMs, and 8 diverse datasets.
\textbf{(3)} We identify a critical trade-off between summary quality and factual faithfulness, demonstrating that increased reasoning complexity does not guarantee better summarization.
\textbf{(4)} We provide practical guidance: explicit reasoning strategies (SC, IR) are optimal for zero-shot summary quality, while {GPT-5} performs best in scenarios demanding rigorous factual consistency.

    





\section{Related Work}

\paragraph{Reasoning in Large Language Models.} 
Reasoning has been studied as a way to improve LLMs~\cite{xu2025towards}. Most approaches focus on explicit reasoning chains that help models think step by step~\citep{wei2022chain, tot_2023, decomposed_prompting_2023}. Prior research has mainly targeted structured tasks with clear intermediate steps, such as solving math problems, performing symbolic reasoning, or answering questions~\citep{wang2022self, diao2024active}. These methods help models break down complex problems, reason through each step, and verify their own conclusions. However, reasoning for open-ended text generation, especially summarization, remains much less explored~\citep{meta_review_logic_2024}.

\paragraph{Reasoning in Summarization.}
Recent work has begun to shift reasoning methods toward abstractive summarization. Some techniques focus on entity- or structure-based reasoning, like Chain-of-Density~\citep{adams2023sparse} and Hierarchical Decomposition~\citep{chang2023booookscore}. Others use reflective reasoning, such as Iterative Refinement, which allows models to review and improve their outputs~\citep{zhang2023summit,yun2025refeedmultidimensionalsummarizationrefinement,zhang2025systematic}. However, most earlier studies test only a few reasoning methods and rely on a narrow range of datasets.

Building on this foundation, our work conducts the first systematic evaluation of a range of reasoning methods. We include both previously studied methods, CoT~\citep{wei2022chain,adams2023sparse,zhang2024comprehensive}, IR~\citep{zhang2023summit}, E2A~\citep{zhang2023extractive} and Deco~\citep{chang2023booookscore,chen2025cothssum}, as well as newly adapted ones such as Cite~\citep{ou2025context}, Plan~\citep{sun2024pearl}, QAG~\citep{sinha2025qa} and SC~\citep{wang2022self,chen2023universal}. 

\section{Method}
\label{sec:method}
\begin{figure*}[ht!]
\centering
\includegraphics[width=\textwidth]{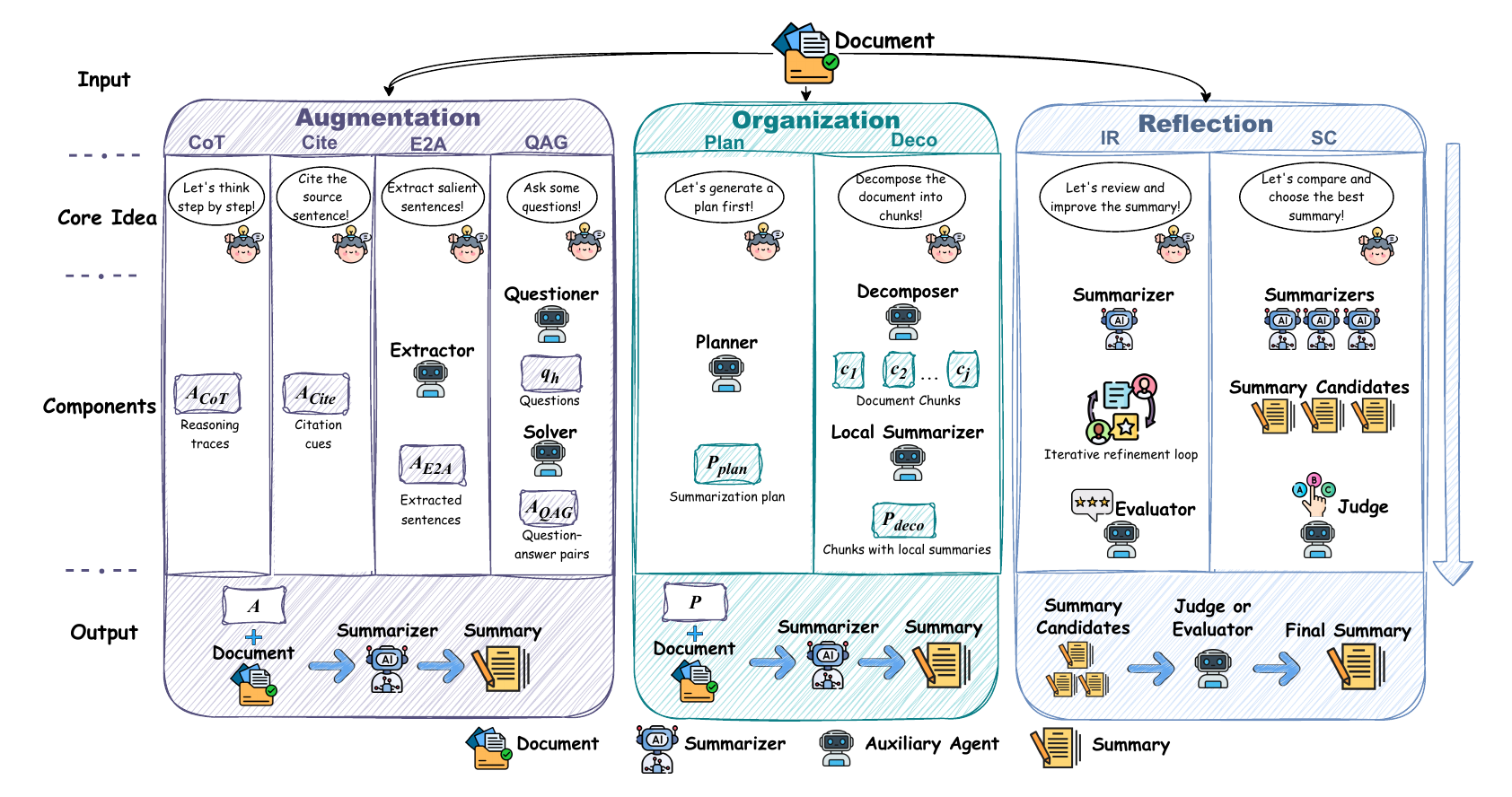}
\caption{
Overview of the 8 reasoning prompting strategies within our framework. We categorize them into three schemes: \textit{Augmentation} expands the input context with auxiliary information; \textit{Organization} structures the generation process via planning; and \textit{Reflection} refines the output through self-evaluation and selection.
}
\phantomsection
\label{fig:all_methods} 
\end{figure*}

Reasoning can be either \emph{implicit and in-model} (e.g., in LRMs) or \emph{explicit and prompt-induced}, which employs intermediate reasoning cues. 
As shown in Figure~\ref{fig:all_methods}, we categorize explicit reasoning methods into three schemes:
(1) \emph{augmentation-based reasoning};
(2) \emph{organization-based reasoning}; and
(3) \emph{reflective reasoning}. 
We describe each scheme in detail below.

\newcommand{\model}{M}            
\newcommand{\doc}{d}              
\newcommand{\summary}{s}          
\newcommand{\context}{C}          
\newcommand{\explain}{e}          
\newcommand{\reason}{z}           
\newcommand{\templ}{\tau}         
\newcommand{\prompt}{x}           

\newcommand{\gen}{\mathcal{G}}  
\newcommand{\augment}{\mathcal{A}}

\subsection{Problem Formulation}

Abstractive summarization is formulated as generating a summary \( s \) from an input document \( d \) using the language model \( \mathcal{M} \). 
The decoding objective can be written as:
\begin{equation}
\hat{s} = \arg\max_{s} \, p_{\mathcal{M}}(s \mid d),
\end{equation}
where \( \hat{s} \) denotes the generated summary. 
We represent the summarization process as:
\begin{equation}
\hat{s} = \mathcal{M}_{\text{sum}}(d).
\end{equation}

\

In the \textit{Vanilla} setting, the model receives a brief instruction along with the document \( d \) and generates the summary in a single pass, i.e., \(\hat{s} = \mathcal{M}_{\text{sum}}(d)\). 
\textit{LRMs} (e.g., \textsc{o1}, \textsc{o3}, \textsc{GPT-5}) use the same instruction, but rely on implicit, in-model reasoning.


\subsection{Augmentation-based Reasoning (CoT, E2A, QAG, Cite).}
In this scheme, the model is guided by the \emph{augmentation} ($A$) that expands the input with intermediate representations, such as stepwise reasoning cues, extracted evidence, or question–answer pairs. We define this generation process as:
\begin{equation}
\hat{s} = \mathcal{M}_{\text{sum}}(d; A),
\end{equation}
where $A$ can be obtained by the following methods:

\textbf{CoT (Chain-of-Thought).}
We adopt the CoT reasoning pattern to summarization~\cite{wei2022chain,adams2023sparse,zhang2024comprehensive}. 
The augmentation \( A_{\text{CoT}} \) provides stepwise reasoning cues, directly embedded in the input prompt. 
The model generates the summary in a single pass:
\begin{equation}
\hat{s} = \mathcal{M}_{\text{sum}}(d; A_{\text{CoT}}).
\end{equation}

\textbf{E2A (Extract-to-Abstract).}
We employ a two-stage extract–abstract pipeline~\cite{dou2021gsum,zhang2023extractive}. 
First, the model extracts ${K}$ salient sentences from the document to form the augmentation \( A_{\text{E2A}} \):
\begin{equation}
A_{\text{E2A}} = \{e_k\}_{k=1}^{K} = \mathcal{M}_{\text{extract}}(d),
\end{equation}
where \( e_k \) denotes each extracted sentence. Next, the model generates the final summary with this additional context:
\begin{equation}
\hat{s} = \mathcal{M}_{\text{sum}}(d; A_{\text{E2A}}).
\end{equation}

\textbf{QAG (Question–Answer Guided).}
Prior works~\cite{durmus2020feqa,liu2024summequal,sinha2025qa} have shown that expert-designed questions can help models improve summarization faithfulness. We design a three-stage procedure that automatically generates questions.
First, the model generates \( H \) document-specific questions:
\begin{equation}
\{q_h\}_{h=1}^{H} = \mathcal{M}_{\text{question}}(d).
\end{equation}
Next, a new model answers each question:
\begin{equation}
a_h = \mathcal{M}_{\text{answer}}(d, q_h).
\end{equation}
Finally, the summary is generated conditioned on the question–answer pairs, which form the augmentation \( A_{\text{QAG}} \) to generate the final summary:
\begin{equation}
A_{\text{QAG}} = \{(q_h, a_h)\}_{h=1}^{H}, \quad
\hat{s} = \mathcal{M}_{\text{sum}}(d; A_{\text{QAG}}).
\end{equation}
The number of questions \( H \) is decided by the question model according to the document length.

\textbf{Cite (Cited Summarization).}
Inspired by cited summarization~\cite{ou2025context}, we propose a single-stage, retrieval-augmented setup.  The augmentation \( A_{\text{Cite}} \) is a set of \( N \) citation units from the document: $A_{\text{Cite}} = \{c_i\}_{i=1}^{N}.$ Each citation unit $c_i$ serves as a referenceable sentence. The model is prompted to produce both the summary \( \hat{s} \) and a set of evidence alignments \( \mathcal{R} \):
\begin{equation}
(\hat{s},\, \mathcal{R}) = \mathcal{M}_{\text{cite}}(d; A_{\text{Cite}}).
\end{equation}
Here, \( \mathcal{R} \) links each summary sentence to its supporting citations in \( A_{\text{Cite}} \).

\subsection{Organization-based Reasoning (Deco, Plan)}
This reasoning scheme structures the summarization process through organization. We denote the organizational guidance prompt as \( P \). 
The model generates the summary as:
\begin{equation}
\hat{s} = \mathcal{M}_{\text{sum}}(d; P),
\end{equation}
where \( P \) encodes the inferred structure or an explicit plan outlining the summary’s organization.

\textbf{Deco (Decomposition).}
We implement a hierarchical decomposition strategy designed for long-document summarization~\cite{chang2023booookscore}. 
The process has three stages. First, the model segments the document \( d \) into \( J \) coherent chunks:
\begin{equation}
\{c_j\}_{j=1}^{J} = \mathcal{M}_{\text{seg}}(d).
\end{equation}
Second, the model generates a local summary \( u_j \) for each chunk:
\begin{equation}
u_j = \mathcal{M}_{\text{sum}}(c_j).
\end{equation}
Finally, the local summaries are combined into a structural plan \( P_{\text{Deco}} \), which guides the final summarization:
\begin{equation}
P_{\text{Deco}} = \{u_j\}_{j=1}^{J}, \qquad 
\hat{s} = \mathcal{M}_{\text{sum}}(d; P_{\text{Deco}}).
\end{equation}
This method captures local information in each chunk and uses it to improve the global summary.

\textbf{Plan (Plan-then-Write).}
Following previous work~\cite{sun2024pearl}, we propose a plan then write strategy by introducing a two-stage structural framework. First, the model generates a compact plan \( P_{\text{Plan}} \):
\begin{equation}
P_{\text{Plan}} = \mathcal{M}_{\text{plan}}(d),
\end{equation}
where \( P_{\text{Plan}} \) includes elements such as 
\(\{\texttt{domain}, \texttt{goal}, \texttt{style}, \texttt{salient\_info}\}\), 
which outlines the summary’s focus and structure. Second, the model generates the final summary conditioned on this plan:
\begin{equation}
\hat{s} = \mathcal{M}_{\text{sum}}(d; P_{\text{Plan}}).
\end{equation} This method mimics the human outlining process to organize key elements before abstraction.

\begin{table*}[t]
    \centering
    \small
    \resizebox{\textwidth}{!}{
    \begin{tabular}{l c c c c c c c}
        \hline
        \multirow{2}{*}{\textbf{Dataset}} & \multirow{2}{*}{\textbf{Domain}} & \multirow{2}{*}{\textbf{Format}} & \multicolumn{5}{c}{\textbf{Statistics}} \\
        & & & Doc.Token & Sum.Token & Density (\%) & Compression & Coverage (\%) \\
        \hline
        CNN/DM~\cite{hermann2015teaching} & News & SDS & 717 & 57 & 44.18 & 13.51 & 53.50 \\
        SAMSum~\cite{gliwa2019samsum} & Dialogue & SDS & 101 & 20 & 16.68 & 5.18 & 31.62 \\
        Reddit~\cite{kim2019abstractive} & Social Media & SDS & 461 & 31 & 13.33 & 18.90 & 35.54 \\
        WikiHow~\cite{koupaee2018wikihow} & Knowledge Base & SDS & 608 & 65 & 9.68 & 10.98 & 36.29 \\
        ArXiv~\cite{cohan2018discourse} & Scientific Paper & SDS & 6,514 & 168 & 16.15 & 42.24 & 58.67 \\
        Multi-News~\cite{fabbri2019multi} & News & MDS & 2,098 & 272 & 0.23 & 7.69 & 45.67 \\
        BookSum~\cite{kryscinski2021booksum} & Novel & LNS & 3,524 & 290 & 1.67 & 12.49 & 31.12 \\
        SciGen~\cite{moosavi2021scigen} & Scientific Paper & TTS & 217 & 123 & 77.83 & 2.32 & 14.85 \\
        \hline
    \end{tabular}
    }
    \caption{Statistics of the included summarization datasets. The ``Format'' column indicates the task type: SDS = Single-Document Summarization, MDS = Multi-Document Summarization, LNS = Long-form Narrative Summarization, and TTS = Table-to-Text Summarization.}
    \label{tab:dataset_statistics}
\end{table*}

\subsection{Reflective Reasoning (IR, SC)}
This reasoning scheme refines summaries through iterative reflection or self-evaluation. 
It allows the model to assess its own outputs and improve them based on the feedback.  We denote reflective feedback or evaluation cues as \( R \). The process typically involves three steps: 
generating one or more candidate summaries, 
evaluating them using \( R \), 
and then revising or selecting the final summary \( \hat{s} \).

\textbf{IR (Iterative Refine).}
We adopt the evaluate then revise framework from prior work~\cite{zhang2023summit, lin2025se}. 
The model iteratively improves its summary through self-evaluation and targeted revision, with early stopping to prevent over-refinement. 
The process starts with an initial draft:
\begin{equation}
s^{(0)} = \mathcal{M}_{\text{sum}}(d).
\end{equation}
For each iteration \( t \in \{1, \dots, T\} \), the model performs a two-step reflective loop. 
First, an evaluator module generates structured feedback \( R^{(t)}_{\text{IR}} \):
\begin{equation}
R^{(t)}_{\text{IR}} = \mathcal{M}_{\text{evaluate}}(d, s^{(t-1)}).
\end{equation}
Second, a reviser module incorporates this feedback to produce an improved summary:
\begin{equation}
s^{(t)} = \mathcal{M}_{\text{revise}}(d, s^{(t-1)}, R^{(t)}_{\text{IR}}).
\end{equation}
The loop terminates when \( R^{(t)}_{\text{IR}} \) signals \texttt{<STOP>}, or after \( T \) iterations. 
The final summary $\hat{s} = s^{(T)}$ is taken from the last iteration.

\textbf{SC (Self-Consistency).}
Building on self-consistency from majority voting~\cite{wang2022self,chen2023universal,li2024improving}, we propose a rubric-based selection framework for abstractive summarization. 
First, the model samples \( N \) candidate summaries from its conditional distribution, using a non-zero decoding temperature:
\begin{equation}
s^{(q)} \sim p_{\mathcal{M}}(s \mid d),
\end{equation}
where $q \in \{1, \dots, N\}$. Next, an LLM-based judge scores each candidate \( s^{(q)} \) against the document \( d \) using a rubric that evaluates criteria such as faithfulness, coverage, and fluency:
\begin{equation}
\text{score}_q = \mathcal{M}_{\text{judge}}(d, s^{(q)}).
\end{equation}
Finally, the final summary is selected as:
\begin{equation}
q^\star = \arg\max_{q} \text{score}_q .
\end{equation}

where the candidate $q^\star$ has the highest score, and the final summary $ \hat{s} = s^{(q^\star)}$ is obtained.

\section{Experimental Setup\label{sec:experiment}}

\subsection{Datasets}
{\setlength{\parskip}{0pt}%
As shown in Table~\ref{tab:dataset_statistics}, we use an evaluation suite of eight representative datasets covering multiple domains~\cite{chu2025domaino1s}, which are categorized into three groups based on their input structure and task type. Full details on data pre-processing are provided in Appendix~\ref{app:implementation_details}.

\textit{Short-Form Documents:}
(1) \textbf{CNN/DM}~\cite{hermann2015teaching} contains concise news articles paired with human-written highlights. 
(2) \textbf{SAMSum}~\cite{gliwa2019samsum} includes messenger-style dialogues and short abstractive summaries. 
(3) \textbf{Reddit}~\cite{kim2019abstractive} features informal, user-generated posts from the TIFU subreddit. 
(4) \textbf{WikiHow}~\cite{koupaee2018wikihow} provides instructional articles with step-wise summaries.
These datasets show the single-document summarization (SDS) setting, with short inputs (100–700 tokens) and moderate compression ratios.

\textit{Long-Form Documents:}
(5) \textbf{ArXiv}~\cite{cohan2018discourse} consists of long scientific papers and their abstracts. 
(6) \textbf{Multi-News}~\cite{fabbri2019multi} merges multiple related news articles into a single summary, requiring cross-document reasoning. 
(7) \textbf{BookSum}~\cite{kryscinski2021booksum} includes chapters and narratives summarized by humans.
These datasets evaluate multi-document (MDS) and long-form narrative summarization (LNS) capabilities, with inputs up to around 6{,}500 tokens.

\textit{Table-to-Text:}
(8) \textbf{SciGen}~\cite{moosavi2021scigen} pairs scientific tables with descriptive textual summaries, testing structured-to-text generation (TTS).

\begin{figure*}[t]
  \centering
  \begin{subfigure}[t]{0.48\textwidth}
    \centering
    \includegraphics[width=\linewidth]{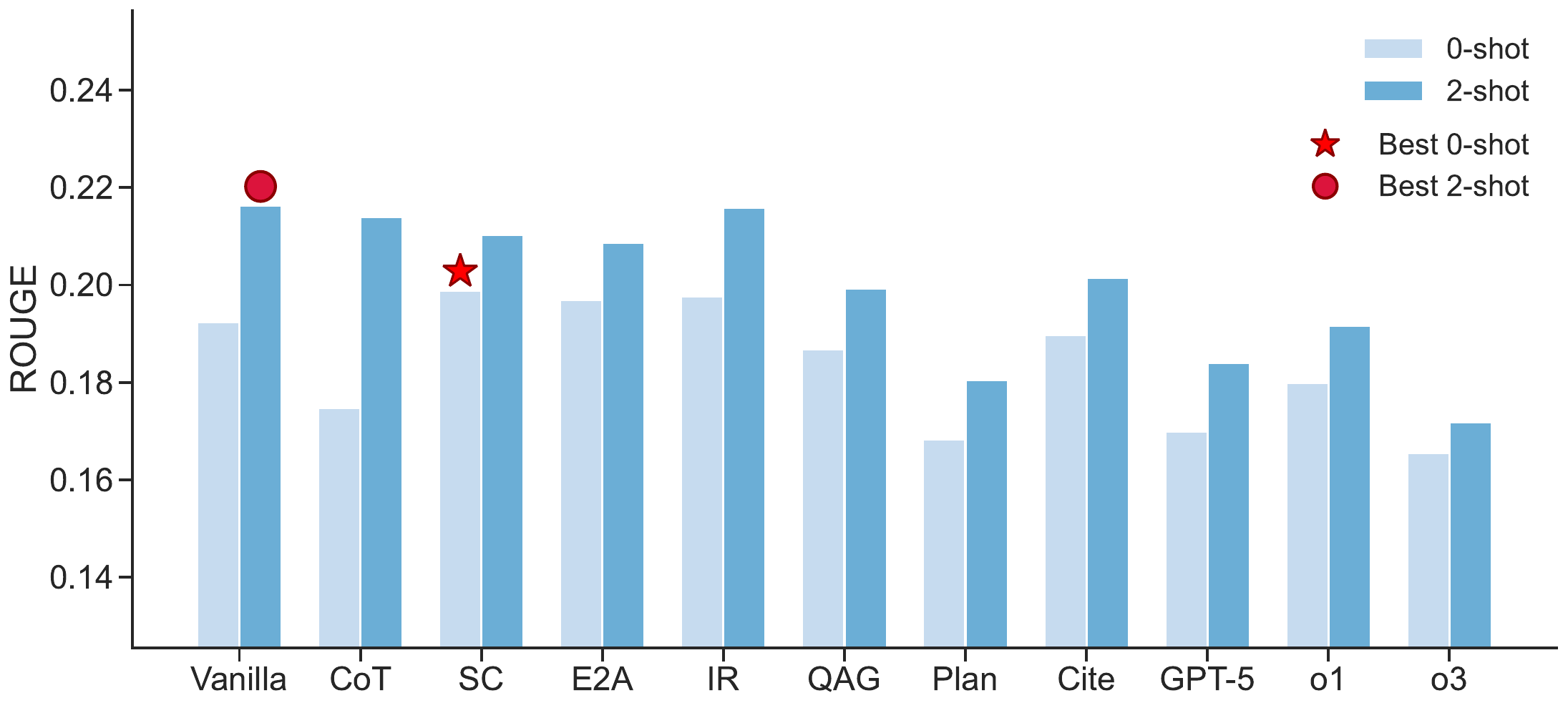}
    \subcaption{ROUGE (average of ROUGE-1/2/L)}
    \label{fig:overall-rouge}
  \end{subfigure}\hfill
  \begin{subfigure}[t]{0.48\textwidth}
    \centering
    \includegraphics[width=\linewidth]{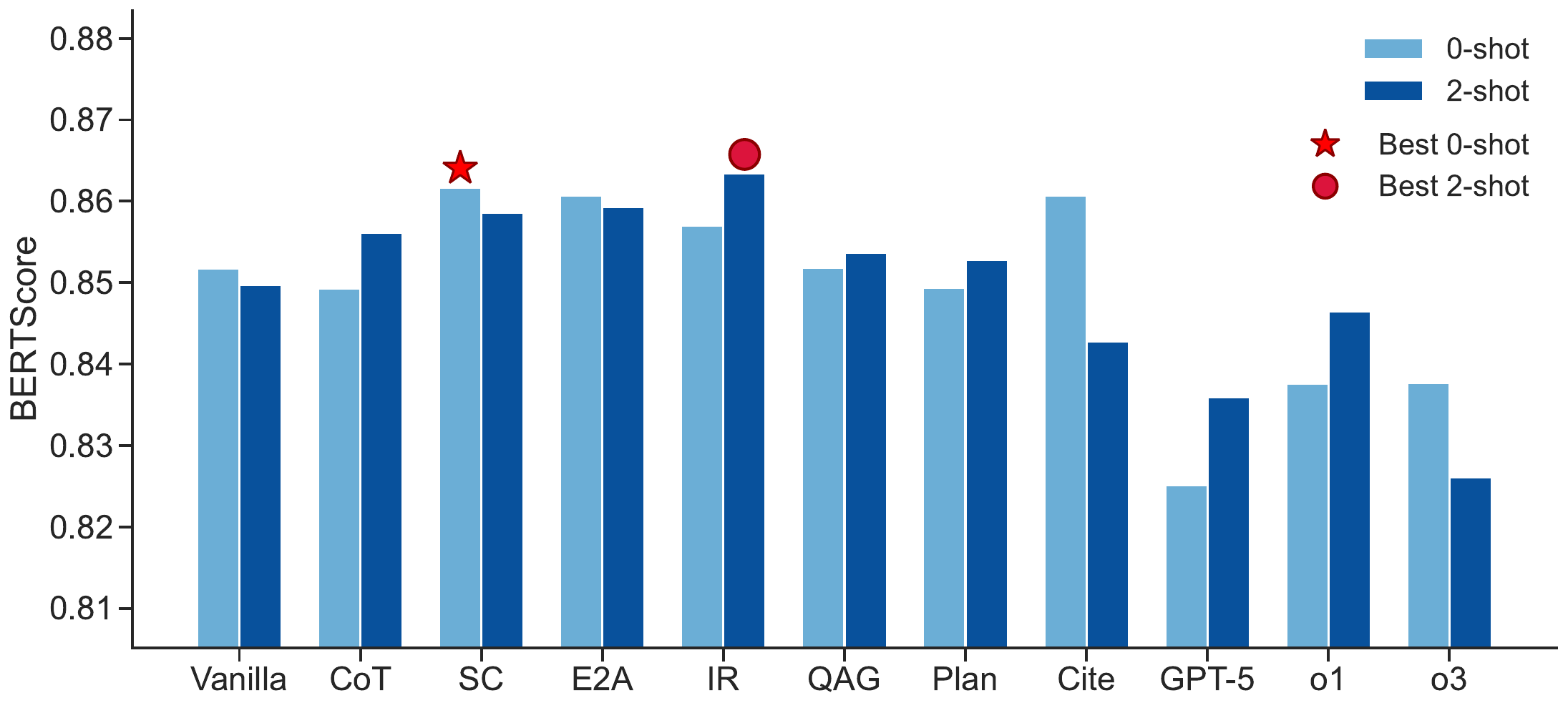}
    \subcaption{BERTScore}
    \label{fig:overall-bertscore}
  \end{subfigure}

  \begin{subfigure}[t]{0.48\textwidth}
    \centering
    \includegraphics[width=\linewidth]{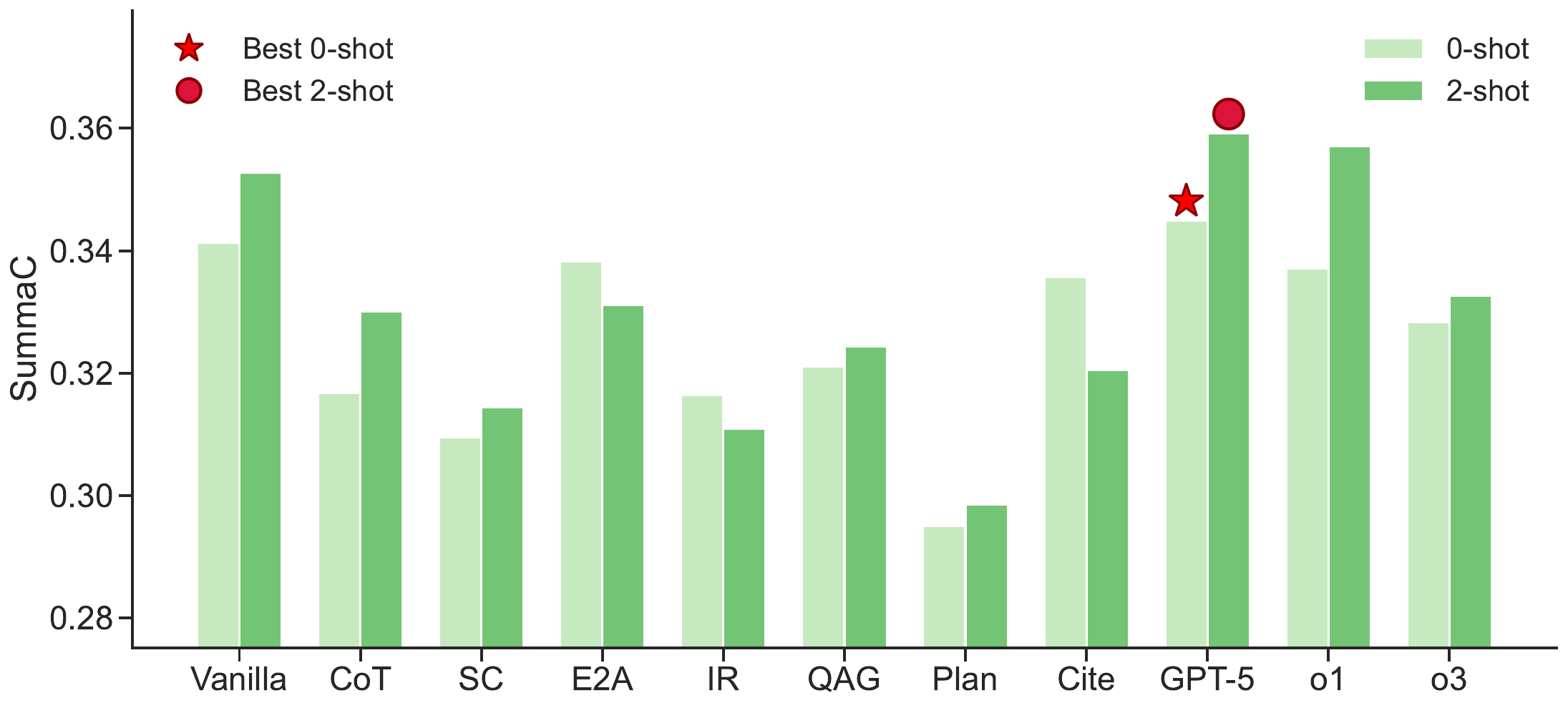}
    \subcaption{SummaC}
    \label{fig:overall-summac}
  \end{subfigure}\hfill
  \begin{subfigure}[t]{0.48\textwidth}
    \centering
    \includegraphics[width=\linewidth]{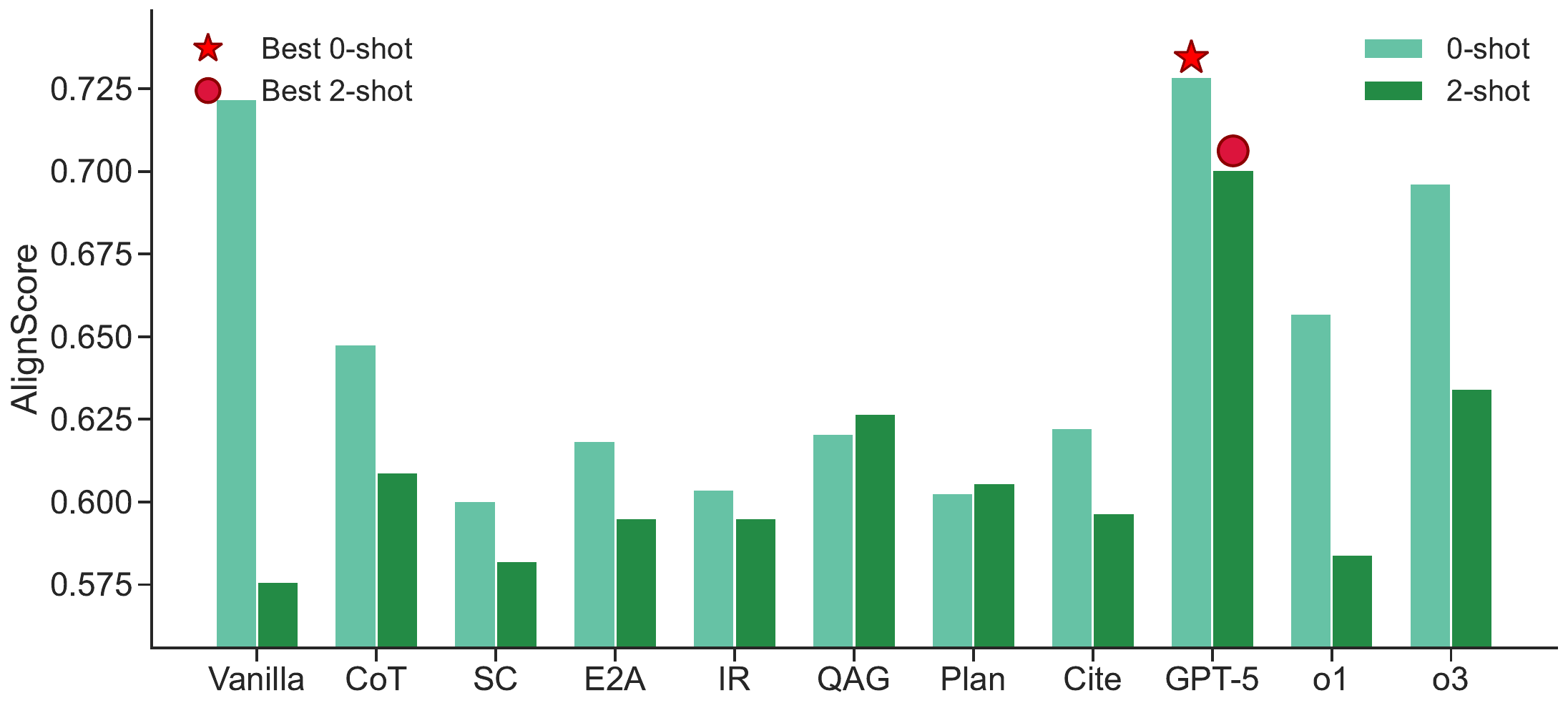}
    \subcaption{AlignScore}
    \label{fig:overall-alignscore}
  \end{subfigure}

  \caption{Summarization performance across 8 datasets. Metrics include reference-based quality (ROUGE, BERTScore) and factual faithfulness (SummaC, AlignScore). Red markers highlight the top methods.}
  \label{fig:overall-metrics}
\end{figure*}

\subsection{Implementation Details}
{\setlength{\parskip}{0pt}%

\paragraph{Models.}
Our experimental evaluation uses the GPT family via Azure AI API\footnote{\url{https://ai.azure.com/}}. We utilize \textsc{GPT-4.1} (1,000K context window) as our main model. We compare its performance against three LRMs: \textsc{o1} (200K context), \textsc{o3} (200K context), and \textsc{GPT-5} (400K context).

\paragraph{Decoding Parameters.}
All configurations run once for each document with a fixed random seed for consistency. We use \texttt{temperature=0} and keep \texttt{top\_p}, \texttt{top\_k}, and other sampling settings at their default values. For \textsc{GPT-4.1}, we set \texttt{max\_output\_tokens=1,000}. For the LRMs, we use \texttt{think\_ability=medium} and \texttt{max\_completion\_tokens=10,000} because these models include internal “thinking” tokens in their completion budget.

\subsection{Evaluation Metrics}
We evaluate summaries using automatic metrics grouped by assessment focus: reference similarity, factual faithfulness, and LLM as a judge.
\textbf{Reference similarity:} 
We report ROUGE (average of ROUGE-1/2/L)~\cite{lin-2004-rouge}, which measures n-gram overlap and longest common subsequence between system and reference summaries. 
For semantic similarity, we use BERTScore F1~\cite{zhang2019bertscore}, which computes contextual embedding similarity between the two summaries.
\textbf{Factual faithfulness:} 
We adopt SummaC~\cite{laban-etal-2022-summac}, which estimates sentence-level consistency via natural language inference, and AlignScore~\cite{zha2023alignscore}, a finetuned metric measuring information alignment between paired texts.
\textbf{LLM as a judge:} 
We use G-Eval~\cite{liu2023geval}, a GPT-based evaluation framework that provides human-aligned scores across three dimensions: Completeness, Conciseness, and Faithfulness. 
G-Eval offers a comprehensive view of overall summary quality beyond traditional overlap-based metrics. Additionally, we conduct a \textbf{human evaluation} using the same rubric as G-Eval, which allows a direct comparison between LLM-based judgments and human assessments.


\section{Results and Analysis}

\subsection{RQ1: Does reasoning actually improve summarization performance compared to standard prompting?}

\paragraph{I. Reasoning helps with summarization only in specific configurations.}
As shown in Figure~\ref{fig:overall-metrics}, we find the impact of reasoning is highly context-dependent. For reference similarity (ROUGE and BERTScore), explicit reasoning methods such as \textit{SC}  and \textit{IR}  achieve the highest scores in the 0-shot setting. However, this advantage disappears in the 2-shot setting, where the \textit{Vanilla} baseline outperforms other methods. For factual faithfulness (SummaC and AlignScore), we observe a different pattern: the LRM \textit{GPT-5} delivers the best results across both settings, suggesting benefits from implicit, model-internal reasoning. Meanwhile, \textit{Vanilla} remains a stable baseline. Overall, reasoning does not always make summarization better. Its benefits depend on the setup. The practical guide on selecting a strategy for each domain is provided in table~\ref{tab:recommended_methods_simplified}.

\paragraph{II. Reasoning-based methods show a quality–faithfulness trade-off.}
Figure~\ref{fig:bs-as-tradeoff} illustrates the relationship between summary quality (BERTScore) and factual faithfulness (AlignScore). 
We observe a negative correlation, with a correlation coefficient of $- 0.685$ and a p-value of $0.014$, indicating that the trade-off is statistically significant.
Most explicit reasoning methods, such as \textit{SC}, \textit{IR}, \textit{QAG}, \textit{Plan}, and \textit{CoT}, group in the high-BERTScore range of about 85 to 86. However, they have a low-to-mid AlignScore of around 60 to 62.
In contrast, the LRMs move towards a lower BERTScore (approximately 82 to 84) but achieve a higher AlignScore. The GPT-5 model sits at the upper-left extreme. Interestingly, the Vanilla baseline is an outlier above the trend line. It achieves both a strong BERTScore and an above-average AlignScore.

\subsection{RQ2: When do reasoning strategies excel across different datasets and context?}

\paragraph{III. SC and IR achieve the best summary quality across most settings.}
As discussed in RQ1, explicit reasoning methods perform best on reference-based metrics. Figures~\ref{fig:overall-rouge} and~\ref{fig:overall-bertscore} show that \textit{SC}  and \textit{IR}  consistently achieve the highest ROUGE and BERTScore in the 0-shot setting. This overall strength reflects complementary, dataset-specific advantages: \textit{SC} leads in short-form summarization (e.g., CNN/DM, SAMSum), while \textit{IR} performs best on long-form datasets (ArXiv, Multi-News). On table-to-text (SciGen), the differences are smaller, with \textit{SC}, \textit{IR}, and \textit{E2A} delivering comparable results. As noted in RQ1, this advantage diminishes in the 2-shot setting, where \textit{Vanilla} becomes highly competitive. In contrast, LRMs (\textit{o1}, \textit{o3}, \textit{GPT-5}) deliver lower scores on reference-based metrics.

\begin{figure}[t]
  \centering
  \includegraphics[width=0.9\columnwidth]{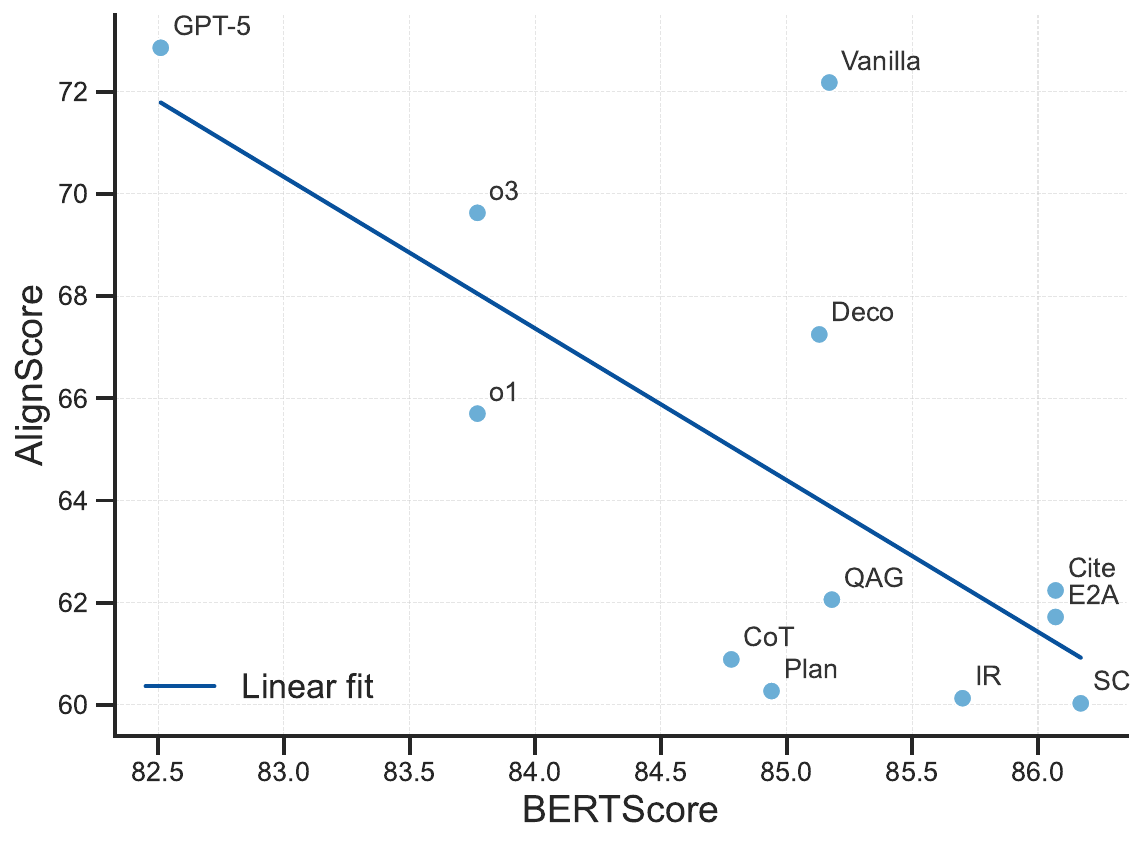}
  \caption{BERTScore vs.\ AlignScore trade-off analysis. The linear fit ($r=-0.685$, $p=0.014$) reveals a significant negative correlation between reference-based quality and factual faithfulness.}
  \label{fig:bs-as-tradeoff}
\end{figure}

\paragraph{IV. LRMs achieve the highest factual faithfulness, especially on long-form summarization.}
 As shown in Figures~\ref{fig:overall-summac} and~\ref{fig:overall-alignscore}, in both 0-shot and 2-shot settings, \textit{GPT-5} achieves the highest average SummaC and AlignScore. Both \textit{o1} and \textit{o3} also do better than explicit reasoning methods. This advantage is clearest in long-form datasets, while \textit{Vanilla} stays competitive only in short-form tasks.

In table-to-text summarization (SciGen), AlignScore approaches 98\% to 100\% across all methods. We infer that factual alignment is easily met because of the structured inputs. Explicit reasoning methods, like SC and IR, lose faithfulness when provided with 2-shot examples. In contrast, LRMs either stay the same or show slight improvement.
Overall, LRMs with implicit reasoning produce the most faithful summaries.

\begin{figure}[t]
  \centering
  \begin{subfigure}{0.485\columnwidth}
    \includegraphics[width=\linewidth]{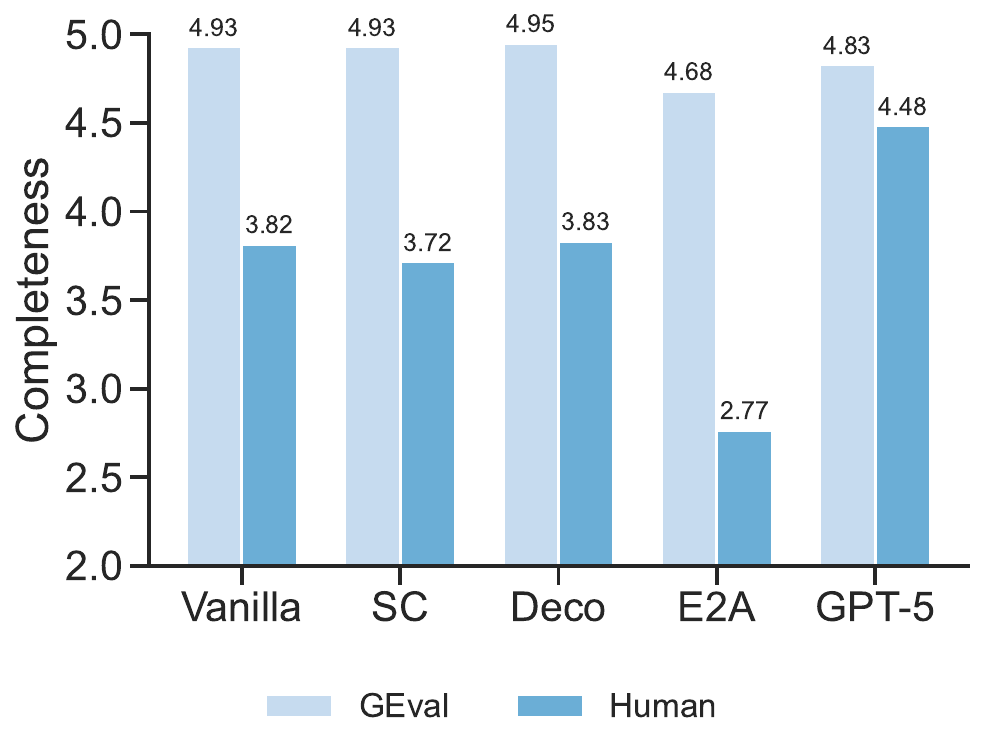}
    \caption{Completeness}
  \end{subfigure}\hfill
  \begin{subfigure}{0.485\columnwidth}
    \includegraphics[width=\linewidth]{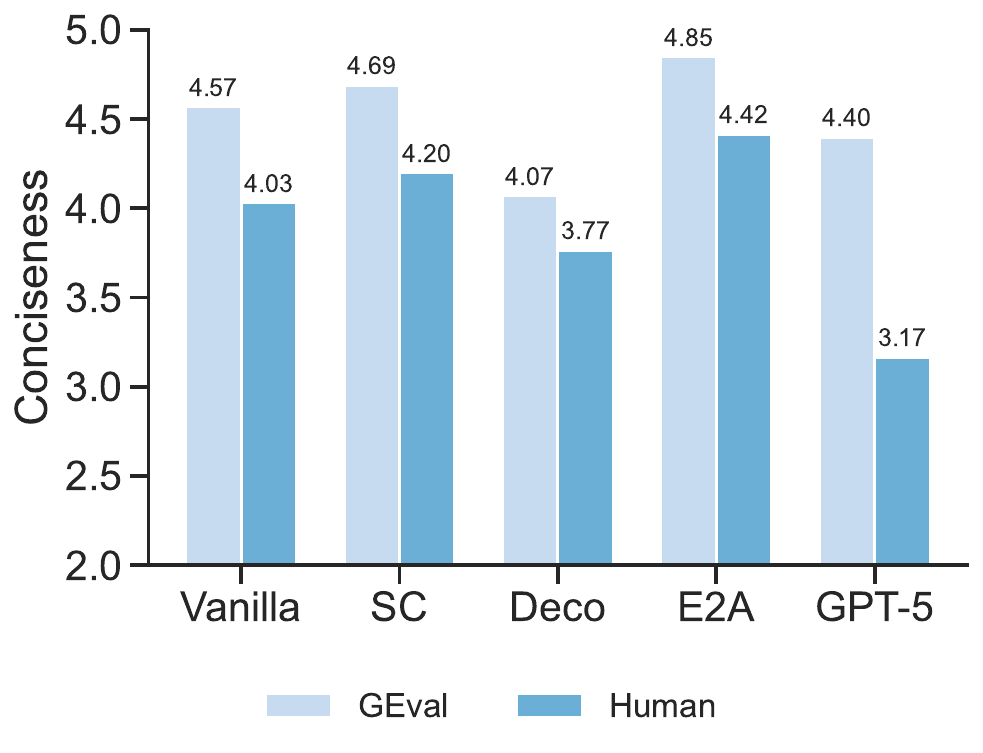}
    \caption{Conciseness}
  \end{subfigure}

  \begin{subfigure}{0.485\columnwidth}
    \includegraphics[width=\linewidth]{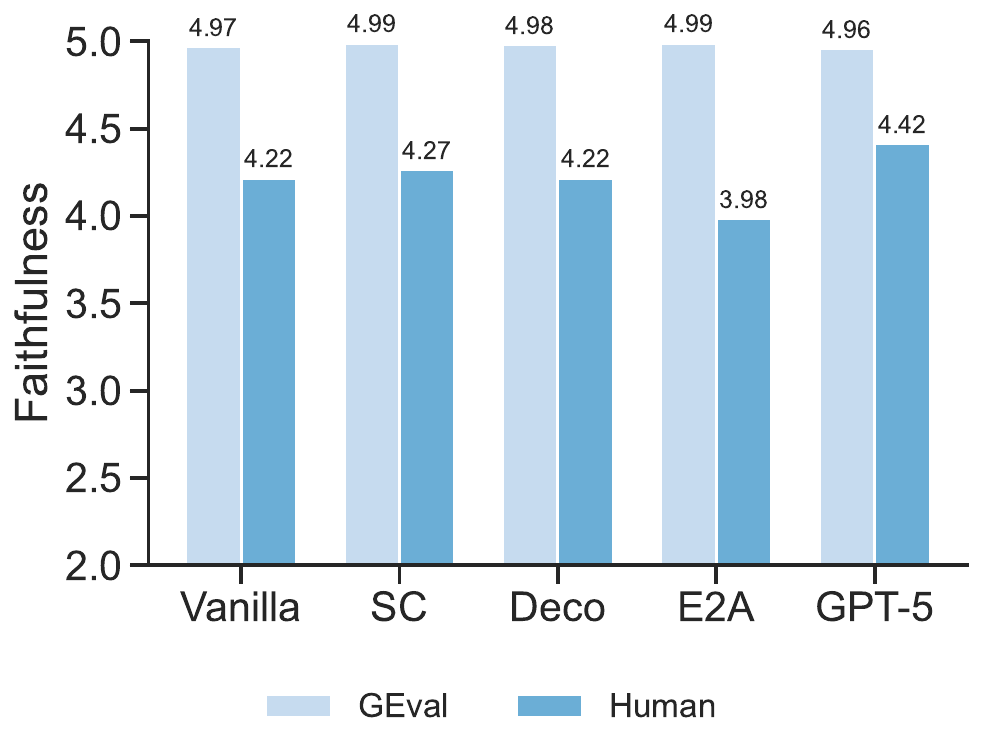}
    \caption{Faithfulness}
  \end{subfigure}\hfill
  \begin{subfigure}{0.485\columnwidth}
    \includegraphics[width=\linewidth]{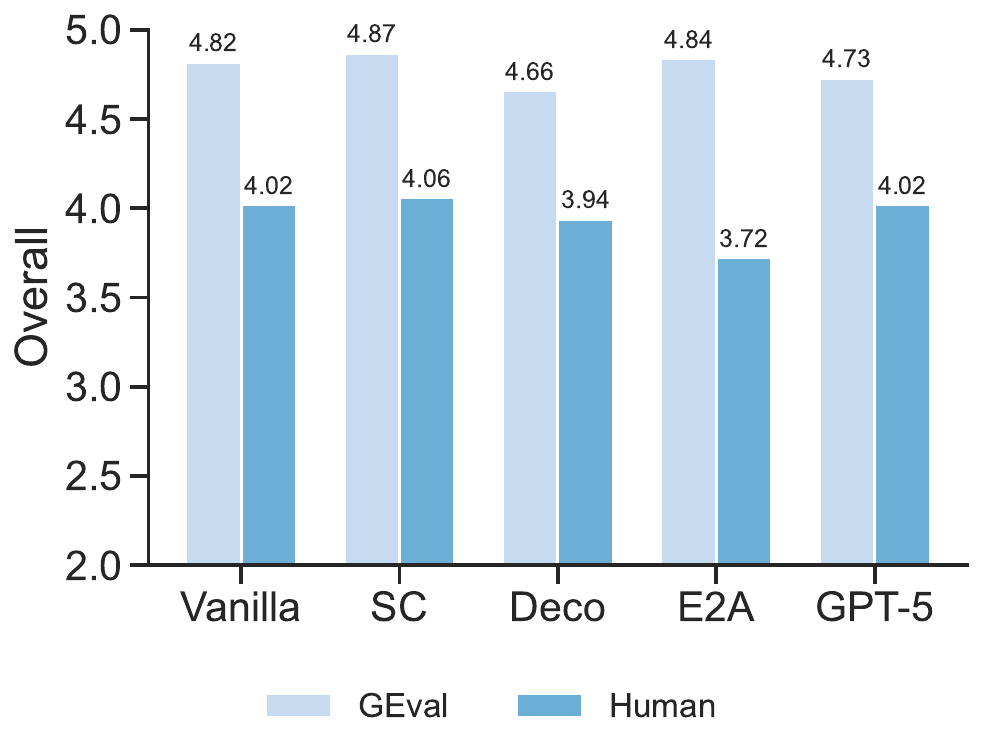}
    \caption{Overall}
  \end{subfigure}

  \caption{Comparison between human evaluation and GEval scores across five systems (Deco, E2A, GPT-5, SC, Vanilla). Each subplot reports one metric. }

  \label{fig:geval-human-2x2}
\end{figure}

\paragraph{V. LLM as a judge overestimates faithfulness compared to humans.}
Following \citet{song2024finesure}, we use GEval to evaluate three metrics: Completeness, Conciseness, and Faithfulness. Overall illustrates the average of the above metrics. As shown in Figure~\ref{fig:geval-human-2x2}, GEval assigns nearly perfect faithfulness scores ($4.96$ to $4.99$) to all systems. In contrast, human ratings are lower and more varied, from $3.98$ to $4.42$ (See Appendix~\ref{Appendix:Human Evaluation} for human evaluation setup).
 GEval considers GPT-5 and E2A nearly the same and even slightly favors E2A ($4.96$ vs.\ $4.99$), whereas humans score GPT-5 much higher in faithfulness ($4.42$ vs.\ $3.98$). Human evaluations also reveal a clear trade-off: GPT-5 is preferred when faithfulness and completeness are most important ($4.42$, $4.48$) but ranks lowest in conciseness ($3.17$), while E2A is the most concise ($4.42$) but loses important content and accuracy ($2.77$ completeness, $3.98$ faithfulness). GEval captures part of this trend by favoring SC and penalizing Deco and GPT-5 for their lack of conciseness. Our results suggest that faithfulness evaluation in summarization cannot rely solely on the LLM.
\begin{table*}[ht!]
\footnotesize
\setlength{\tabcolsep}{3pt}
\renewcommand{\arraystretch}{1.08}
\begin{tabular}{p{1.55cm}p{13.55cm}}
\toprule
\textbf{Document} & 
[...] If you are going to be eating the pieces raw, they can be a little longer, but 
\textcolor{blue}{if you’re cooking with them, $2$--$3$ inches is best}. Each chunk does not have to be the exact same length, just divide the vegetable evenly. The general rule for juliennes is that you’d be able to eat them in one bite, but you can cut them into longer chunks if you want. When cutting juliennes, be careful not to cut yourself as you slice. \textcolor{blue}{Flip the slices} you’ve just cut so that they are piled on top of each other. Then cut them into \textcolor{blue}{sticks that are $\tfrac{1}{8}$\,in $(0.3\,\text{cm})$ to $\tfrac{1}{16}$\,in $(0.15\,\text{cm})$ thick}. The goal with juliennes is to get them all to a fairly uniform size.
\\
\midrule
\textbf{Human Ref.} & 
[...] Cut rounded vegetables into a rectangle. Chop the vegetable into \textcolor{blue}{chunks that are $2$--$3$ inches long}. Slice the vegetable down the length. \textcolor{blue}{Stack 2--3 slices} and cut again.
\\
\midrule
\textbf{Vanilla} & 
[...] Slice vegetables into rectangular chunks, discarding or saving scraps. For julienne cuts, divide vegetables evenly; \textcolor{blue}{raw pieces can be longer, but cooked ones should be $2$--$3$ inches}. Cut slices into \textcolor{blue}{sticks $\tfrac{1}{8}$\,in $(0.3\,\text{cm})$ to $\tfrac{1}{16}$\,in $(0.15\,\text{cm})$ thick}, \textcolor{blue}{stacking slices} for uniformity, and always be careful not to cut yourself. Aim for uniform, bite-sized pieces.
\\
\midrule
\textbf{SC} & 
[...] Cut off stems and bottoms as needed, slice vegetables into rectangular chunks, and either discard or use scraps. For raw pieces, longer cuts are acceptable; \textcolor{blue}{for cooking, $2$--$3$ inch lengths are best}. Aim for uniform, bite-sized \textcolor{blue}{sticks about $\tfrac{1}{8}$ to $\tfrac{1}{16}$ inch thick}, \textcolor{blue}{stacking slices} before cutting, and always be careful not to cut yourself.
\\
\midrule
\textbf{Deco} & 
[...] Prepare vegetables by peeling them if desired, removing unwanted parts like stems, bottoms, or potato eyes, and cutting them into manageable rectangular chunks based on your intended use and preference. For julienning, \textcolor{blue}{stack the slices} and cut them into \textcolor{blue}{uniform sticks about $\tfrac{1}{8}$ to $\tfrac{1}{16}$ inch thick} \textcolor{red}{and $2$--$3$ inches long}, aiming for consistency in size to ensure even cooking and presentation.
\\
\midrule
\textbf{E2A} & 
[...] Whether or not you peel vegetables depends on how you plan to eat them. Slice off one side at a time to create rectangular chunks, aiming for pieces that are generally bite-sized, though longer chunks are acceptable if preferred. \textcolor{red}{The key is to cut the juliennes to a fairly uniform size.}
\\
\midrule
\textbf{GPT-5} & 
[...] 
\newline
- Square the vegetable by slicing off sides to form a rectangular chunk; discard or use trimmings. This is especially helpful for potatoes, zucchini, and squash. \newline
- Cut the chunk into even slabs, \textcolor{blue}{stack them}, then slice into \textcolor{blue}{uniform sticks about $\tfrac{1}{8}$--$\tfrac{1}{16}$ inch ($\approx 0.3$--$0.15$ cm) thick}. Aim for one-bite, fairly uniform pieces; be careful not to cut yourself. \newline
- Length: \textcolor{blue}{for cooking, $2$--$3$ inches is best; raw pieces can be a little longer}. Divide the vegetable into evenly sized chunks.
\\
\bottomrule
\end{tabular}
\caption{Case study from WikiHow comparing system summaries with the source document and a human reference. \textcolor{blue}{Blue text} marks source-faithful procedural details, especially the conditional length constraint and the stacking/cutting steps. \textcolor{red}{Red text} marks errors, including over-generalization in \textbf{Deco} and critical omission in \textbf{E2A}.}
\label{tab:case_study_short}
\end{table*}

\subsection{RQ3: How do different reasoning methods impact summarization results?}

\paragraph{VI. Summary abstractiveness is negatively correlated with reference-based quality and factual faithfulness.}
\begin{figure}[t]
  \centering
  \begin{subfigure}{0.45\columnwidth}
    \includegraphics[width=\linewidth]{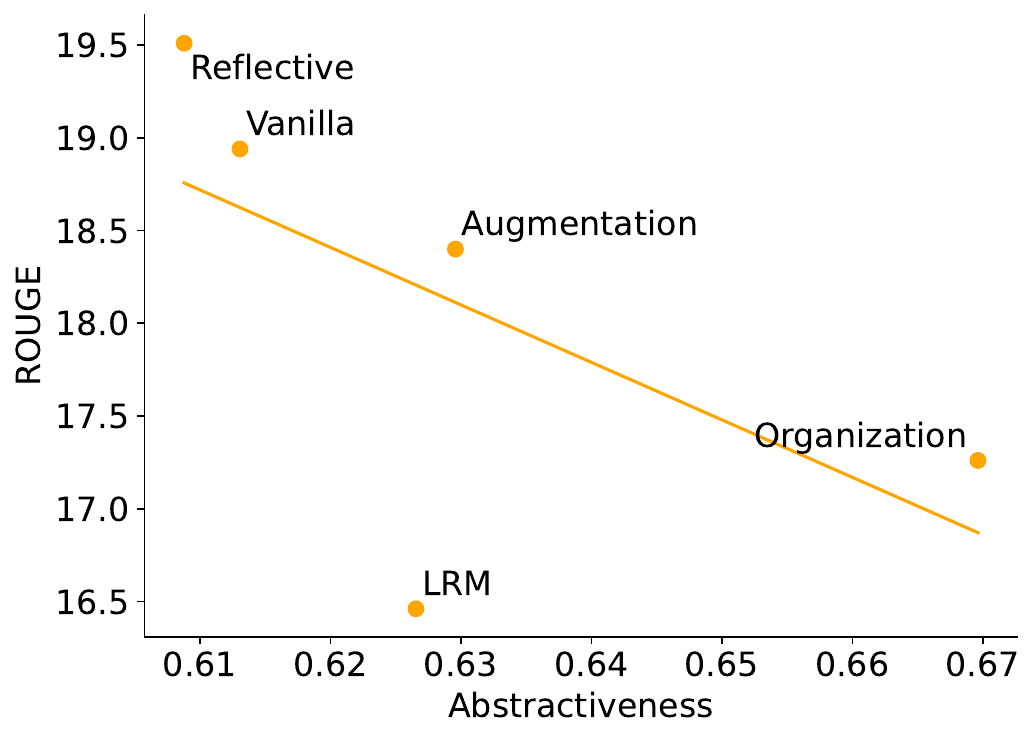}
    \caption{ROUGE}
  \end{subfigure}\hfill
  \begin{subfigure}{0.45\columnwidth}
    \includegraphics[width=\linewidth]{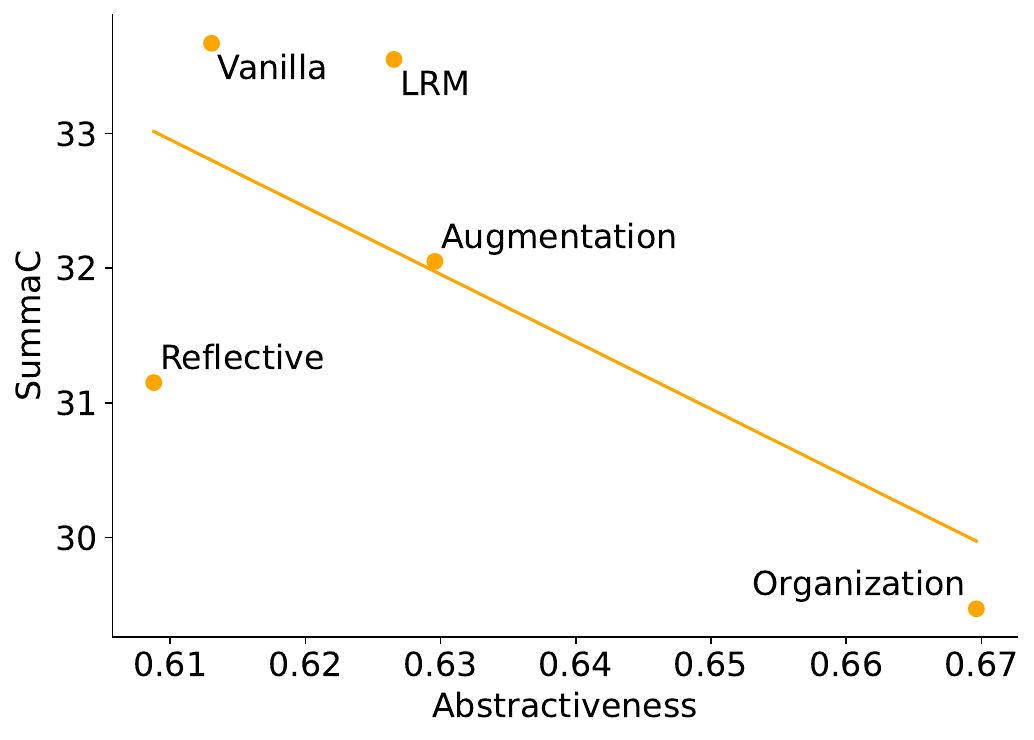}
    \caption{SummaC}
  \end{subfigure}

  \caption{Higher abstractiveness correlates negatively with both reference-based quality (ROUGE) and factual faithfulness (SummaC).}
  \vspace{-5pt}
  \label{fig:abs-tradeoff-2x1}
\end{figure}

\begin{figure}[t]
  \centering
  \begin{subfigure}{0.45\columnwidth}
    \includegraphics[width=\linewidth]{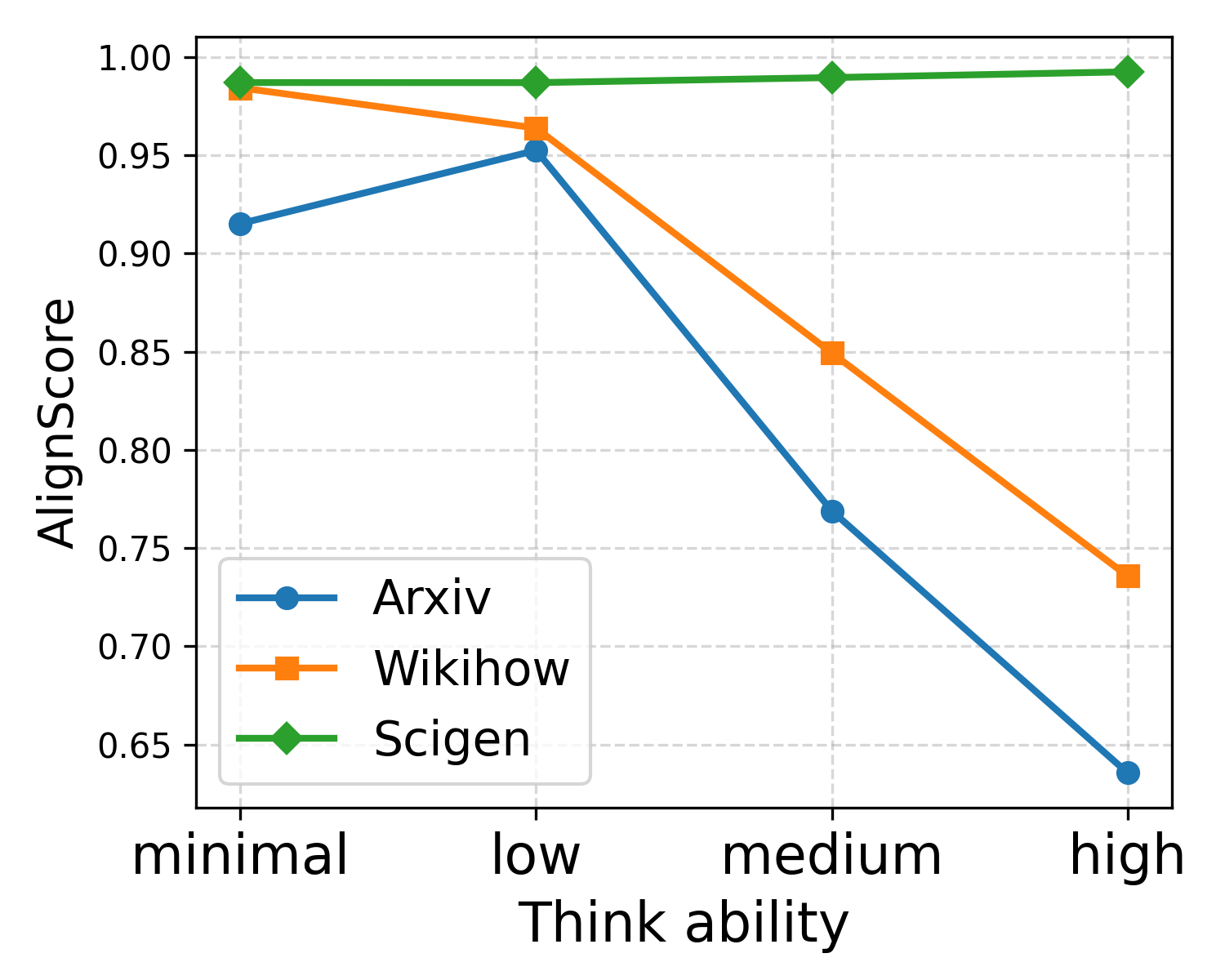}
    \caption{AlignScore}
  \end{subfigure}\hfill
  \begin{subfigure}{0.45\columnwidth}
    \includegraphics[width=\linewidth]{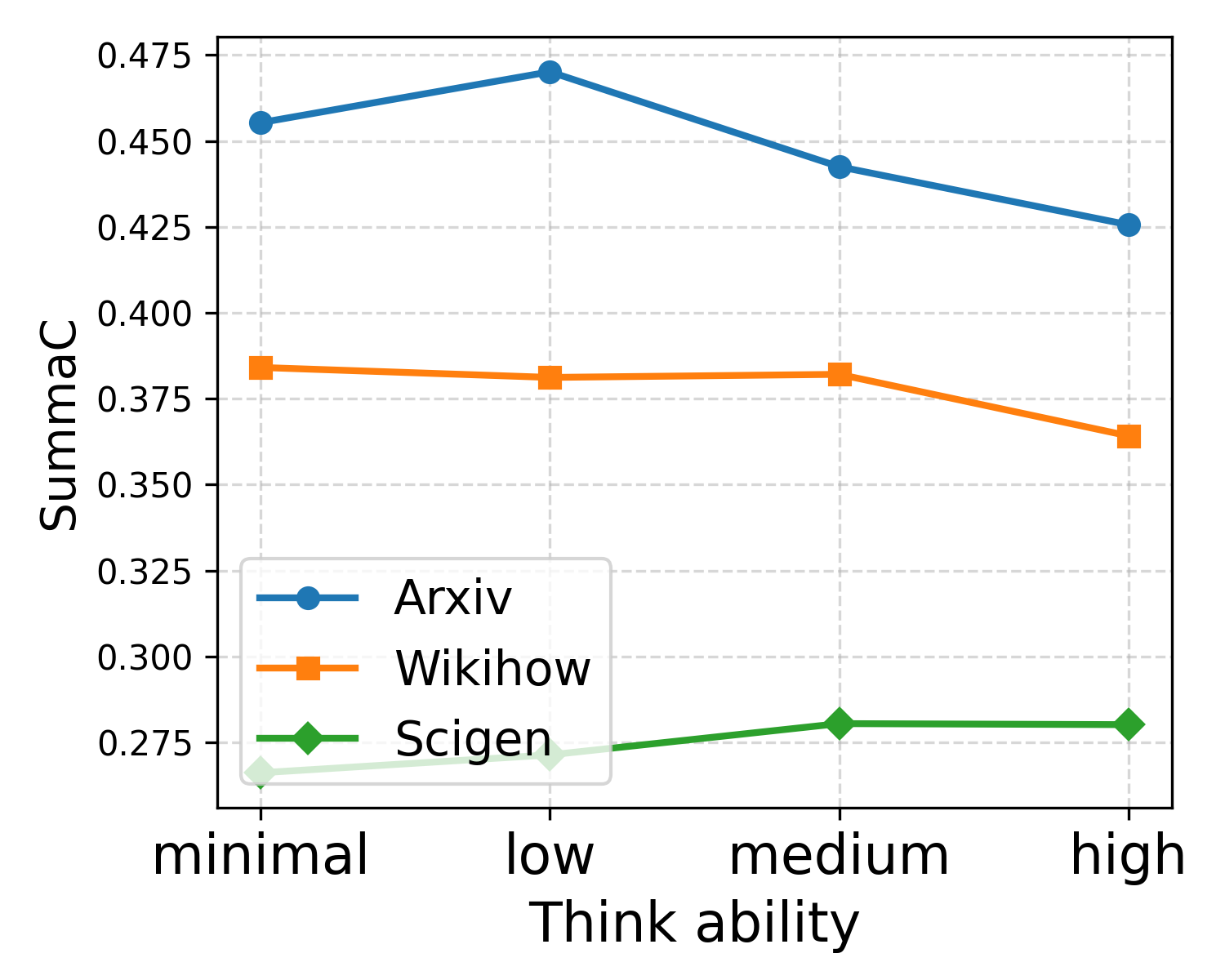}
    \caption{SummaC}
  \end{subfigure}

  \begin{subfigure}{0.45\columnwidth}
    \includegraphics[width=\linewidth]{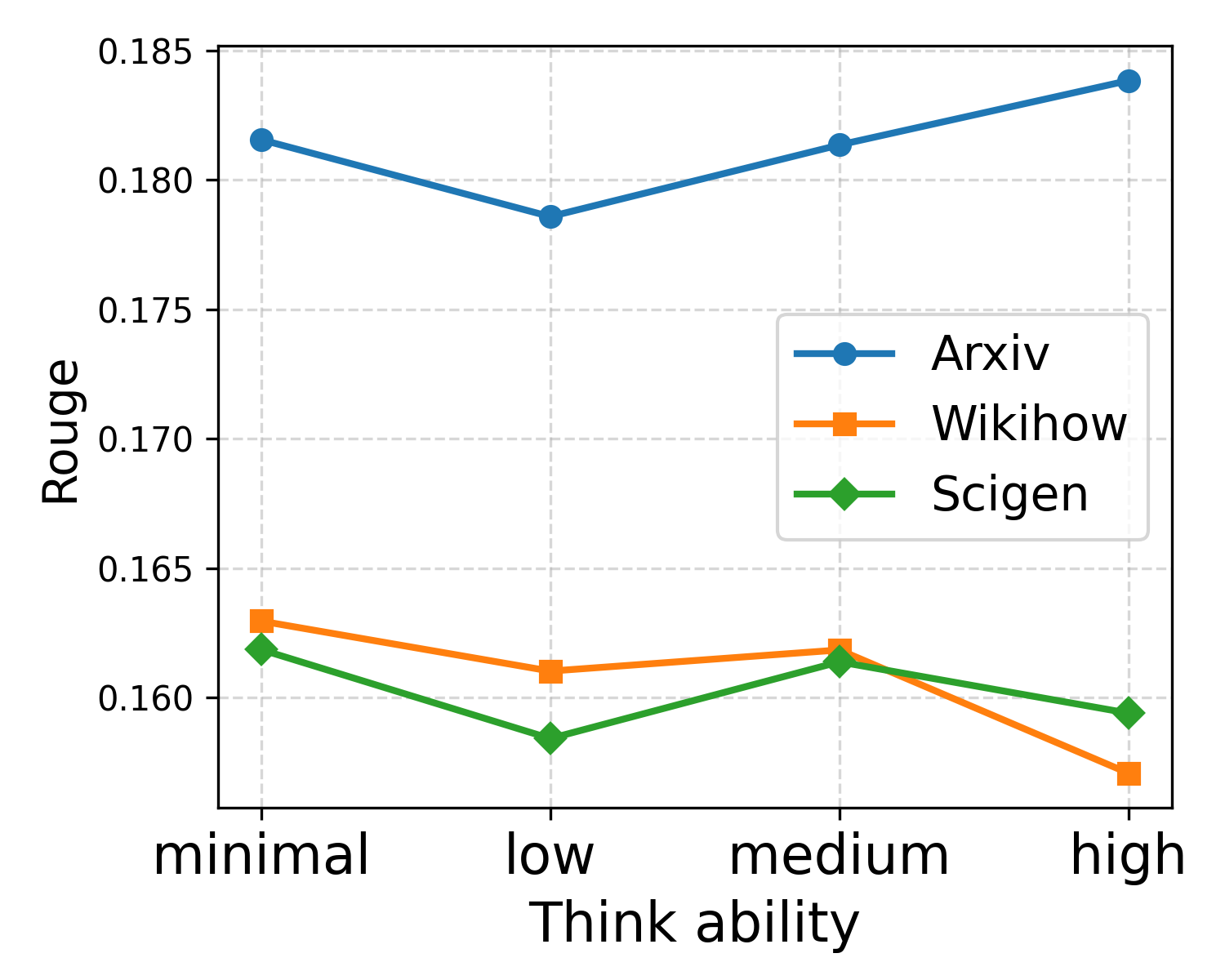}
    \caption{ROUGE}
  \end{subfigure}\hfill
  \begin{subfigure}{0.45\columnwidth}
    \includegraphics[width=\linewidth]{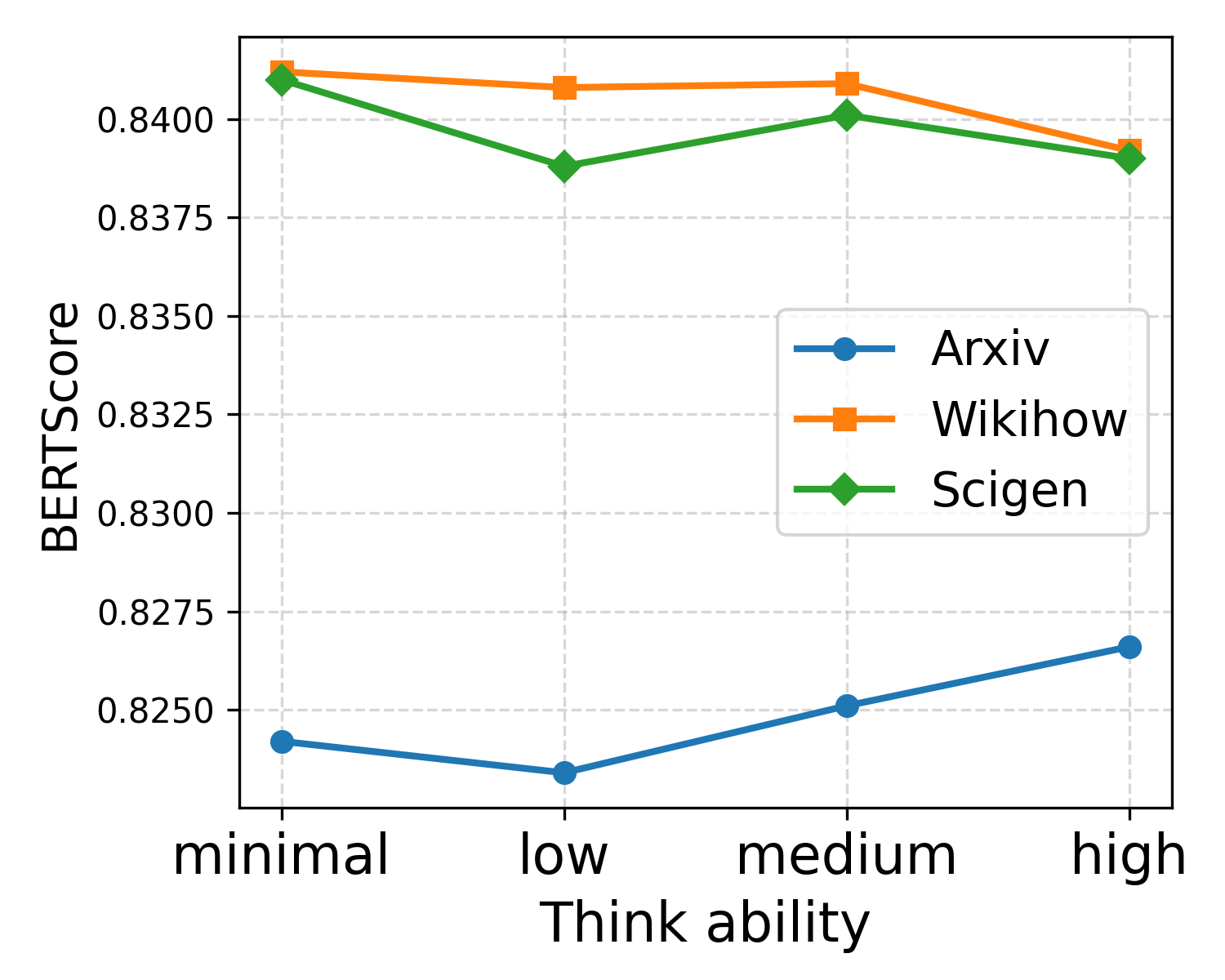}
    \caption{BERTScore}
  \end{subfigure}

  \caption{Effect of GPT-5 think ability (from minimal to high) on 4 evaluation metrics across 3 datasets. }
  \vspace{-5pt}
  \label{fig:think-ability-2x2}
\end{figure}

As outlined in Section~\ref{sec:method}, we group systems into five reasoning paradigms and measure their average abstractiveness~\cite{yuan2025domainsum}. 
In Figure~\ref{fig:abs-tradeoff-2x1}, \textit{Vanilla} and \textit{Reflective} are the least abstract and achieve the highest ROUGE scores. 
\textit{Augmentation} methods fall in the middle on both metrics. 
\textit{Organization} methods are the most abstract and perform the worst on both metrics. 
LRMs break this trend: they are relatively extractive but trade lower ROUGE for higher SummaC. 
This suggests that abstractiveness interacts with model family and internal reasoning style.

We further conduct a qualitative case study to investigate specific model behaviors. As shown in Table~\ref{tab:case_study_short}, Deco's chunk-based aggregation loses global conditional context, causing it to over-generalize a conditional rule and apply the “2–3 inches” cooking length to all julienne sticks, while E2A's extraction bottleneck excludes the final cutting step, resulting in an incomplete procedure. By comparison, Vanilla and GPT-5 largely preserve the original conditional structure. These examples suggest that summarization requires accurate compression instead of creative reasoning. Reasoning pipelines that fragment or filter context risk driving models to fill non-existent logical gaps or omit important information. The complete case study table is provided in Appendix~\ref{app:case_study}.

\paragraph{VII. Risk of Over-Thinking: Increased reasoning reduces faithfulness without improving quality.}
We analyze the effect of a hyperparameter named ``think ability'', which controls the depth of internal reasoning within \textit{GPT-5}. 
We select one representative dataset from each of the three groups: short-form (\textit{WikiHow}), long-form (\textit{ArXiv}), and table-to-text (\textit{SciGen}). 
As shown in Figure~\ref{fig:think-ability-2x2}, factual faithfulness consistently declines as think ability increases: \textit{AlignScore} shows the steepest drop, while \textit{SummaC} exhibits a similar but milder trend. 
In contrast, reference-based quality metrics (\textit{ROUGE} and \textit{BERTScore}) remain stable. This observed drop suggests that the LRM can ``over-think'' the source instead of maintaining faithful compression. 
Overall, too much internal reasoning poses a clear risk. It undermines factual grounding without improving summary quality.

\paragraph{VIII. Controlled validation beyond GPT: More internal reasoning does not always help summarization.}
Our main benchmark focuses on the GPT family. To test whether this pattern also appears beyond GPT, we conduct an additional controlled study on \textit{Gemini 2.5 Flash}. Specifically, we vary its internal \emph{thinking budget} (\texttt{tb}) while keeping the summarization prompt fixed to \textit{Vanilla}. We use the same evaluation protocol and metrics as in Section~\ref{sec:experiment}. Table~\ref{tab:gemini_budget_ablation} reports results on three representative datasets: short-form \textit{CNN/DM}, long-form \textit{ArXiv}, and table-to-text \textit{SciGen}.

We find that increasing the thinking budget does not consistently improve summarization performance. \textit{BERTScore} stays nearly unchanged across different budgets. For ROUGE, the clearest pattern appears on \textit{ArXiv}, where performance drops from $0.2284$ at \texttt{tb=0} to $0.1954$ at \texttt{tb=1024}. On \textit{CNN/DM} and \textit{SciGen}, the best ROUGE is obtained at the middle setting \texttt{tb=256}, rather than at the largest budget. The faithfulness metrics show a similarly mixed pattern across datasets. Overall, these results support our main conclusion that internal reasoning is highly context-dependent in summarization, and that simply allocating more reasoning budget does not reliably lead to better summaries.

\nobreak\smallskip
\begin{minipage}{\columnwidth}
\centering
\setlength{\tabcolsep}{3pt}
\renewcommand{\arraystretch}{1.06}
\scriptsize

\textbf{(a) Reference similarity.}\\[0.3em]
\resizebox{0.94\columnwidth}{!}{%
\begin{tabular}{l c c c c}
\hline
\textbf{Metric} & \textbf{tb} & \textbf{CNN/DM} & \textbf{ArXiv} & \textbf{SciGen} \\
\hline
\multirow{3}{*}{ROUGE}
  & 0    & 0.2374 & \textbf{0.2284} & 0.1737 \\
  & 256  & \textbf{0.2423} & 0.2278 & \textbf{0.1769} \\
  & 1024 & 0.2366 & 0.1954 & 0.1716 \\
\hline
\multirow{3}{*}{BERTScore}
  & 0    & 0.8713 & 0.8399 & 0.8467 \\
  & 256  & 0.8712 & \textbf{0.8410} & 0.8489 \\
  & 1024 & \textbf{0.8721} & 0.8402 & \textbf{0.8501} \\
\hline
\end{tabular}%
}

\vspace{0.55em}

\textbf{(b) Factual faithfulness.}\\[0.3em]
\resizebox{0.94\columnwidth}{!}{%
\begin{tabular}{l c c c c}
\hline
\textbf{Metric} & \textbf{tb} & \textbf{CNN/DM} & \textbf{ArXiv} & \textbf{SciGen} \\
\hline
\multirow{3}{*}{SummaC}
  & 0    & 0.3224 & 0.4831 & \textbf{0.3074} \\
  & 256  & \textbf{0.3275} & 0.4756 & 0.2941 \\
  & 1024 & 0.3180 & \textbf{0.5317} & 0.2957 \\
\hline
\multirow{3}{*}{AlignScore}
  & 0    & 0.4916 & 0.8779 & 0.9886 \\
  & 256  & \textbf{0.6156} & \textbf{0.9761} & \textbf{0.9940} \\
  & 1024 & 0.5280 & 0.7472 & 0.9871 \\
\hline
\end{tabular}%
}

\vspace{0.25em}
\begingroup
\captionsetup{font=footnotesize,skip=3pt,justification=raggedright,singlelinecheck=false}
\captionof{table}{Gemini 2.5 Flash with different thinking budgets (\texttt{tb}) under the \textit{Vanilla} prompt. Bold marks the best \texttt{tb} for each dataset and metric.}
\label{tab:gemini_budget_ablation}
\endgroup
\end{minipage}
\smallskip

\section{Conclusion}
This paper presents the first large-scale comparative study of reasoning paradigms in LLM-based summarization. 
We evaluate eight prompting strategies and three LRMs across eight diverse datasets. 
Our findings challenge the assumption that reasoning always improves summarization. 
We uncover a trade-off: explicit prompting enhances quality but reduces faithfulness, whereas implicit reasoning in LRMs generally shows the opposite. 
Crucially, we find that excessive internal reasoning triggers creative gap-filling that harms factual consistency. 
Overall, our results suggest that reasoning in summarization is highly context-dependent: stronger reasoning does not always lead to better summaries, and more complex reasoning can reduce faithfulness without improving final summary quality.

\section*{Acknowledgement}
This work was supported in part by National Science Foundation awards 2149133, 2004014, and 2138259, as well as by computing resources
provided through the NSF ACCESS projects CIS250940 and CIS24061.

\newpage
\section*{Limitations}

In our evaluation, we used 100-instance subsets per dataset rather than the full test sets. This choice was made to support a broad comparison across eight diverse datasets, eight reasoning paradigms, and three LRMs. Previous research has similarly adopted sampled evaluation for large-scale LLM studies~\cite{goyal2023newssummarizationevaluationera, zhang-etal-2024-benchmarking}. While this allows for extensive cross-domain analysis, future work may further verify our findings on full test splits.

Our study mainly focuses on English summarization. Although the selected datasets span diverse domains, including news, dialogue, social media, structured table-to-text~\cite{guo2025sqlforge}, and scientific papers, extending our analysis to multilingual settings or specialized vertical domains (e.g., legal~\cite{zhang2025citalaw}, biomedical~\cite{yuan2024structure,  ou2025experienceretrievalaugmentationelectronichealth}, or clinical~\cite{shen-etal-2026-patient}) remains an important direction for future research.

We conduct our experiments using the GPT model family (GPT-4.1, o1, o3, GPT-5). This ensures a consistent baseline capability, allowing us to isolate the impact of different reasoning strategies without confounding factors from varying model architectures or training distributions. However, relying on closed-source APIs restricts fine-grained inspection of internal reasoning processes (e.g., attention weights). Consequently, our analysis of "internal reasoning in the LRM" relies on observing output behaviors rather than internal mechanisms~\cite{lin2026mmfinereason, zhu2026dissectingfailuredynamicslarge}. Verifying these patterns on open-weight models~\cite{li2025curriculum} would be a valuable extension.

 Regarding evaluation, we employ a combination of human and automated metrics. While human evaluation serves to validate the trends observed in automated scores, scaling manual annotation to the full extent of such a large benchmark remains challenging. Furthermore, we do not conduct a fine-grained latency analysis. Some resource-heavy methods (e.g., QAG, SC, IR) need multiple generations, which may inherently trade inference cost for quality. Optimizing this efficiency trade-off~\cite{nie2026attnpo,liu2026chartverse} is a practical consideration for deployment that warrants future engineering-focused studies.

\newpage

\bibliography{custom}

\appendix
\FloatBarrier
\setlength{\textfloatsep}{8pt plus 2pt minus 2pt}
\setlength{\floatsep}{8pt plus 2pt minus 2pt}
\setlength{\intextsep}{8pt plus 2pt minus 2pt}
\captionsetup{font=small,skip=4pt}

\newpage

\section{Detailed Implementation Setup}
\label{app:implementation_details}

\subsection{Dataset Sampling and Reproducibility}
Conducting a large-scale evaluation involving computationally intensive Large Reasoning Models (LRMs) such as o1 and GPT-5, alongside multi-step reasoning strategies (e.g., Iterative Refinement), presents significant cost and latency challenges. To manage these constraints while ensuring statistically meaningful comparisons, we performed our evaluation on a fixed subset of the data.

Specifically, for each of the eight datasets we used, we randomly sampled $100$ instances from the official test split. This sampling process was controlled by a unified global random seed (\texttt{seed=42}) to ensure reproducibility. Consequently, our final evaluation suite consists of 800 distinct document-summary pairs.  

\subsection{Software and Environment}
All experiments were conducted using Python. We utilized the standard HuggingFace \texttt{datasets} library for data loading and processing. The random sampling was implemented using Python's native \texttt{random} module initialized with the aforementioned seed. All model interactions were performed via the Azure OpenAI API, with generation parameters fixed (temperature=0) as detailed in Section 4.2 to ensure deterministic outputs.

\section{Detailed Main Results}

We report detailed results for all methods and datasets, with ROUGE, BERTScore, SummaC, and AlignScore
scores across 0-shot and 2-shot settings. Tables~\ref{tab:coverage_results_8ds_oneblock},
\ref{tab:coverage_results_8ds_2shot}, \ref{tab:faithfulness_results_8ds_sumc_as}, and
\ref{tab:faithfulness_results_8ds_sumc_as_2shot} present these full results.

\section{Performance Results Grouped by Datasets}

Here we reorganize the results by dataset to better understand how reasoning strategies
behave across short-form, long-form, multi-document, and table-to-text summarization.
As shown in Tables~\ref{tab:rouge_bs_grouped_small}, \ref{tab:rouge_bs_grouped_2shot},
\ref{tab:faithfulness_grouped_0shot}, and \ref{tab:faithfulness_grouped_2shot}, this
grouped view highlights domain-specific behavior more clearly.

\begin{table}[htbp]
\setlength{\tabcolsep}{3pt}
\renewcommand{\arraystretch}{1.1}
\centering
\tiny

\begin{tabular}{l|cc|cc|cc|cc}
\hline
& \multicolumn{2}{c|}{\textbf{Short}} & \multicolumn{2}{c|}{\textbf{Long}} & \multicolumn{2}{c|}{\textbf{Table}} & \multicolumn{2}{c}{\textbf{Avg}} \\
\textbf{System} & ROUGE & BS & ROUGE & BS & ROUGE & BS & ROUGE & BS \\
\hline
\rowcolor{gray!15}
\multicolumn{9}{c}{\textbf{\textsl{LLM (GPT-4.1)}}} \\
\hline
Vanilla & 19.43 & 85.29 & 19.42 & 84.98 & 18.60 & 85.22 & 19.24 & 85.17 \\
COT & 17.00 & 85.27 & 18.40 & 84.56 & 17.63 & 84.16 & 17.61 & 84.78 \\
SC & 20.12 & 86.85 & 20.09 & 85.37 & 19.12 & 86.01 & 19.89 & 86.17 \\
Cite & 19.41 & 86.77 & 19.06 & 85.23 & 17.95 & 85.93 & 18.97 & 86.07 \\
Deco & 16.55 & 85.28 & 18.44 & 84.75 & 17.70 & 85.39 & 17.43 & 85.13 \\
QAG & 17.78 & 85.36 & 20.15 & 85.04 & 18.27 & 85.04 & 18.68 & 85.18 \\
Plan & 16.34 & 85.22 & 17.50 & 84.55 & 16.80 & 84.98 & 16.83 & 84.94 \\
IR & 19.42 & 85.97 & 20.68 & 85.44 & 19.10 & 85.52 & 19.77 & 85.70 \\
E2A & 20.11 & 86.77 & 19.74 & 85.24 & 18.76 & 85.92 & 19.69 & 86.07 \\
\hline
\rowcolor{gray!15}
\multicolumn{9}{c}{\textbf{\textsl{Large Reasoning Models (LRM)}}} \\
\hline
o1 & 18.35 & 84.04 & 18.41 & 84.27 & 16.60 & 82.46 & 17.98 & 83.77 \\
o3 & 16.61 & 83.74 & 17.43 & 83.85 & 15.08 & 83.70 & 16.54 & 83.77 \\
GPT-5 & 17.34 & 81.66 & 17.66 & 83.47 & 15.26 & 82.75 & 16.99 & 82.51 \\
\hline
\end{tabular}
\caption{Zero-shot ROUGE (R) and BS scores grouped by dataset type: Short, Long, and Table. Avg = per-system mean.}
\label{tab:rouge_bs_grouped_small}
\end{table}

\begin{table}[ht]
\setlength{\tabcolsep}{3pt}
\renewcommand{\arraystretch}{1.1}
\centering
\tiny

\begin{tabular}{l|cc|cc|cc|cc}
    \hline
    & \multicolumn{2}{c|}{\textbf{Short}}
    & \multicolumn{2}{c|}{\textbf{Long}}
    & \multicolumn{2}{c|}{\textbf{Table}}
    & \multicolumn{2}{c}{\textbf{Avg}} \\
    \textbf{System}
        & ROUGE & BS
        & ROUGE & BS
        & ROUGE & BS
        & ROUGE & BS \\
    \hline
    \rowcolor{gray!15}
    \multicolumn{9}{c}{\textbf{\textsl{LLM (GPT-4.1)}}} \\
    \hline
    Vanilla & 22.56 & 85.57 & 20.95 & 83.81 & 19.96 & 86.05 & 21.63 & 84.97 \\
    COT     & 23.05 & 85.89 & 21.33 & 85.40 & 19.45 & 85.11 & 21.95 & 85.61 \\
    SC      & 21.65 & 86.24 & 20.83 & 85.26 & 19.13 & 86.16 & 21.03 & 85.86 \\
    Cite    & 20.67 & 83.95 & 19.69 & 84.03 & 19.43 & 86.39 & 20.15 & 84.28 \\
    Deco    & 18.64 & 84.48 & 18.16 & 82.44 & 18.84 & 85.78 & 18.49 & 83.88 \\
    QA      & 19.87 & 85.83 & 20.41 & 84.88 & 18.64 & 85.03 & 19.92 & 85.37 \\
    Plan    & 18.00 & 85.92 & 17.97 & 84.44 & 18.43 & 85.28 & 18.04 & 85.28 \\
    IR      & 22.18 & 86.88 & 21.32 & 85.62 & 19.94 & 86.34 & 21.58 & 86.34 \\
    E2A     & 21.79 & 86.24 & 20.06 & 85.42 & 19.60 & 86.27 & 20.87 & 85.93 \\
    \hline
    \rowcolor{gray!15}
    \multicolumn{9}{c}{\textbf{\textsl{Large Reasoning Models (LRM)}}} \\
    \hline
    o1      & 20.07 & 84.84 & 18.47 & 84.11 & 17.66 & 85.50 & 19.17 & 84.65 \\
    o3      & 17.45 & 81.99 & 17.37 & 82.88 & 15.51 & 84.28 & 17.18 & 82.61 \\
    GPT-5   & 19.07 & 83.53 & 18.36 & 83.60 & 15.86 & 83.83 & 18.40 & 83.59 \\
    \hline
\end{tabular}
\caption{2-shot ROUGE (R) and BS scores grouped by dataset type: Short, Long, and Table. Avg = per-system mean.}
\label{tab:rouge_bs_grouped_2shot}
\end{table}

\begin{table}[ht]
\setlength{\tabcolsep}{3pt}
\renewcommand{\arraystretch}{1.1}
\centering
\tiny

\begin{tabular}{l|cc|cc|cc|cc}
    \hline
    & \multicolumn{2}{c|}{\textbf{Short}}
    & \multicolumn{2}{c|}{\textbf{Long}}
    & \multicolumn{2}{c|}{\textbf{Table}}
    & \multicolumn{2}{c}{\textbf{Avg}} \\
    \textbf{System}
        & SumC & AS
        & SumC & AS
        & SumC & AS
        & SumC & AS \\
    \hline
    \rowcolor{gray!15}
    \multicolumn{9}{c}{\textbf{\textsl{LLM (GPT-4.1)}}} \\
    \hline
    Vanilla & 39.69 & 60.55 & 26.48 & 80.49 & 34.83 & 93.78 & 34.13 & 72.18 \\
    COT     & 29.98 & 68.35 & 38.64 & 38.09 & 28.22 & 99.45 & 33.01 & 60.89 \\
    SC      & 35.82 & 38.40 & 23.73 & 79.27 & 33.19 & 88.79 & 30.96 & 60.03 \\
    Cite    & 38.35 & 41.69 & 27.34 & 81.60 & 33.10 & 86.38 & 33.57 & 62.24\\
    Deco    & 36.06 & 55.35 & 23.97 & 80.70 & 29.31 & 76.49 & 30.68 & 67.25 \\
    QAG     & 28.84 & 65.48 & 38.90 & 45.32 & 24.84 & 98.63 & 32.11 & 62.06 \\
    Plan    & 34.08 & 39.01 & 23.15 & 80.73 & 30.28 & 83.90 & 29.51 & 60.27\\
    IR      & 36.00 & 38.85 & 25.52 & 80.86 & 32.63 & 85.05 & 31.65 & 60.13 \\
    E2A     & 30.73 & 68.82 & 41.05 & 40.78 & 24.59 & 97.13 & 33.83 & 61.72 \\
    \hline
    \rowcolor{gray!15}
    \multicolumn{9}{c}{\textbf{\textsl{Large Reasoning Models (LRM)}}} \\
    \hline
    o1      & 29.89 & 68.92 & 40.36 & 50.86 & 29.09 & 97.30 & 33.71 & 65.70 \\
    o3      & 38.35 & 58.74 & 24.65 & 78.39 & 35.34 & 86.93 & 32.83 & 69.63 \\
    GPT-5   & 38.97 & 61.01 & 27.39 & 82.02 & 37.91 & 92.73 & 34.50 & 72.86 \\
    \hline
\end{tabular}
\caption{Zero-shot Faithfulness grouped by dataset type: Short, Long, and Table. Metrics: SummaC (SumC) and AlignScore (AS). Avg = per-system mean.}
\label{tab:faithfulness_grouped_0shot}
\end{table}

\begin{table}[ht]
\setlength{\tabcolsep}{3pt}
\renewcommand{\arraystretch}{1.1}
\centering
\tiny

\begin{tabular}{l|cc|cc|cc|cc}
    \hline
    & \multicolumn{2}{c|}{\textbf{Short}}
    & \multicolumn{2}{c|}{\textbf{Long}}
    & \multicolumn{2}{c|}{\textbf{Table}}
    & \multicolumn{2}{c}{\textbf{Avg}} \\
    \textbf{System}
        & SumC & AS
        & SumC & AS
        & SumC & AS
        & SumC & AS \\
    \hline
    \rowcolor{gray!15}
    \multicolumn{9}{c}{\textbf{\textsl{LLM (GPT-4.1)}}} \\
    \hline
    Vanilla & 33.81 & 62.66 & 39.79 & 37.17 & 27.59 & 98.50 & 35.27 & 57.58 \\
    COT     & 29.98 & 68.35 & 38.64 & 38.09 & 28.22 & 99.45 & 33.01 & 60.89 \\
    SC      & 28.11 & 63.41 & 38.16 & 37.90 & 24.61 & 98.44 & 31.44 & 58.22 \\
    Cite    & 28.77 & 66.94 & 38.88 & 37.24 & 24.66 & 97.84 & 32.05 & 59.66 \\
    Deco    & 26.62 & 67.48 & 37.16 & 42.38 & 25.04 & 98.59 & 30.37 & 61.95 \\
    QA-G    & 29.69 & 66.36 & 37.06 & 45.66 & 25.62 & 98.91 & 31.95 & 62.67 \\
    Plan    & 26.58 & 66.06 & 36.36 & 40.30 & 23.46 & 99.45 & 29.85 & 60.57 \\
    IR      & 28.02 & 66.35 & 37.41 & 37.33 & 24.47 & 98.70 & 31.10 & 59.51 \\
    E2A     & 30.10 & 65.06 & 39.35 & 39.04 & 26.46 & 98.71 & 33.11& 59.51 \\
    \hline
    \rowcolor{gray!15}
    \multicolumn{9}{c}{\textbf{\textsl{Large Reasoning Models (LRM)}}} \\
    \hline
    o1      & 30.42 & 70.50 & 39.78 & 47.62 & 26.46 & 97.95 & 32.44 & 65.35 \\
    o3      & 28.74 & 54.95 & 42.45 & 63.10 & 23.84 & 98.25 & 33.27 & 63.42 \\
    GPT-5   & 33.93 & 61.53 & 41.70 & 71.61 & 26.53 & 99.39 & 35.92 & 70.04 \\
    \hline
\end{tabular}
\caption{2-shot Faithfulness grouped by dataset type: Short, Long, and Table. Metrics: SummaC (SumC) and AlignScore (AS). Avg = per-system mean.}
\label{tab:faithfulness_grouped_2shot}
\end{table}

\newcommand{\DSI}{CNN/DM}
\newcommand{\DSII}{SAMSum}
\newcommand{\DSIII}{Reddit}
\newcommand{\DSIV}{WikiHow}
\newcommand{\DSV}{ArXiv}
\newcommand{\DSVI}{Multi-News}
\newcommand{\DSVII}{BookSum}
\newcommand{\DSVIII}{SciGen}

\newcommand{\metricheadcov}{\textbf{ROUGE} & \textbf{BS}}


\begin{table*}[ht]
\setlength{\tabcolsep}{3pt} 
\renewcommand{\arraystretch}{1.2}
\centering
\scriptsize 

\resizebox{\textwidth}{!}{
\begin{tabular}{l|cc|cc|cc|cc|cc|cc|cc|cc|cc}
    \hline
    & \multicolumn{2}{c|}{\textbf{\DSI}}
    & \multicolumn{2}{c|}{\textbf{\DSII}}
    & \multicolumn{2}{c|}{\textbf{\DSIII}}
    & \multicolumn{2}{c|}{\textbf{\DSIV}}
    & \multicolumn{2}{c|}{\textbf{\DSV}}
    & \multicolumn{2}{c|}{\textbf{\DSVI}}
    & \multicolumn{2}{c|}{\textbf{\DSVII}}
    & \multicolumn{2}{c|}{\textbf{\DSVIII}}
    & \multicolumn{2}{c}{\textbf{Average}} \\
    \textbf{System} 
        & \metricheadcov & \metricheadcov & \metricheadcov & \metricheadcov 
        & \metricheadcov & \metricheadcov & \metricheadcov & \metricheadcov
        & \metricheadcov \\
    \hline
    \rowcolor{gray!15}
    \multicolumn{19}{c}{\textbf{\textsl{LLM (GPT-4.1)}}} \\
    \hline
    Vanilla  & 21.27 & 84.24 
             & 26.88 & 88.85 
             & 12.82 & 84.22 
             & 16.75 & 83.85 
             & 20.97 & 83.76 
             & 19.79 & 85.41 
             & 17.49 & 85.76 
             & 17.96 & 85.28 
             & 19.24 & 85.17 \\

    COT      & 20.88 & 84.88 
         & 21.03 & 88.11 
         & 11.36 & 83.85 
         & 14.72 & 84.22 
         & 20.39 & 83.59 
         & 19.95 & 85.44 
         & 14.87 & 84.65 
         & 16.55 & 84.65 
         & 17.47 & 84.92\\

    SC       & 22.23 & 86.94 
             & 26.67 & 90.01 
             & 14.82 & 85.55 
             & 16.75 & 84.89 
             & 23.12 & 84.84 
             & 19.47 & 85.50 
             & 17.68 & 85.76 
             & 18.35 & 85.85 
             & \cellcolor{red!24}19.89 & \cellcolor{red!24} 86.17 \\

    Cite     & 22.76 & 87.16 
             & 24.25 & 89.51 
             & 13.04 & 85.11 
             & 17.58 & 85.30 
             & 22.95 & 84.79 
             & 17.89 & 85.26 
             & 16.34 & 85.64 
             & 16.93 & 85.79 
             & 18.97 &\cellcolor{red!16} 86.07 \\

    Deco     & 19.47 & 86.47 
             & 21.62 & 86.85 
             & 10.58 & 83.39 
             & 14.51 & 84.40 
             & 19.91 & 83.51 
             & 18.26 & 85.09 
             & 17.15 & 85.66 
             & 17.96 & 85.65 
             & 17.43 & 85.13 \\

    QAG      & 20.03 & 84.60
         & 24.78 & 89.30
         & 11.48 & 83.44
         & 14.83 & 84.12
         & 22.23 & 84.20
         & 20.72 & 85.44
         & 17.49 & 85.47
         & 17.87 & 84.89
         & 18.68 & 85.18 \\

    Plan     & 18.94 & 84.70 
             & 19.57 & 88.39 
             & 12.43 & 83.65 
             & 14.43 & 84.15 
             & 19.76 & 83.87 
             & 18.33 & 85.09 
             & 14.40 & 84.69 
             & 16.76 & 85.01 
             & 16.83 & 84.94 \\

    IR       & 22.68 & 86.16 
             & 25.10 & 88.79 
             & 13.68 & 84.27 
             & 16.22 & 84.67 
             & 23.65 & 84.85 
             & 20.67 & 85.64 
             & 17.71 & 85.84 
             & 18.42 & 85.34 
             &\cellcolor{red!16} 19.77 &\cellcolor{red!08} 85.70 \\

    E2A      & 22.86 & 87.01 
             & 26.51 & 89.85 
             & 13.87 & 85.36 
             & 17.21 & 84.85 
             & 22.68 & 84.52 
             & 19.05 & 85.43 
             & 17.49 & 85.77 
             & 17.84 & 85.78 
             &\cellcolor{red!08} 19.69 & \cellcolor{red!16}86.07 \\

    \hline
    \rowcolor{gray!15}
    \multicolumn{19}{c}{\textbf{\textsl{Large Reasoning Models (LRM)}}} \\
    \hline
    o1       & 21.12 & 85.01 
             & 23.50 & 87.59 
             & 12.74 & 80.74 
             & 16.04 & 82.82 
             & 20.15 & 83.95 
             & 17.97 & 84.17 
             & 17.12 & 84.69 
             & 15.21 & 81.16 
             & 17.98 & 83.77 \\

    o3       & 18.54 & 82.54 
             & 22.02 & 87.44 
             & 11.01 & 81.03 
             & 14.87 & 83.96 
             & 16.61 & 81.71 
             & 17.99 & 84.67 
             & 17.68 & 85.18 
             & 13.62 & 83.64 
             & 16.54 & 83.77 \\

    GPT-5    & 19.64 & 82.41 
             & 23.52 & 84.60 
             & 12.45 & 80.19 
             & 13.75 & 79.46 
             & 16.79 & 81.98 
             & 18.81 & 83.67 
             & 17.39 & 84.75 
             & 13.54 & 82.99 
             & 16.99 & 82.51 \\

    \hline
\end{tabular}}
\caption{Summary Coverage across 8 datasets with ROUGE (average of ROUGE 1/2/L) and BERTScore (BS). The last two columns show per-system averages across all 8 datasets.}
\label{tab:coverage_results_8ds_oneblock}
\end{table*}

\begin{table*}[ht]
\setlength{\tabcolsep}{3pt}
\renewcommand{\arraystretch}{1.2}
\centering
\scriptsize

\resizebox{\textwidth}{!}{
\begin{tabular}{l|cc|cc|cc|cc|cc|cc|cc|cc|cc}
    \hline
    & \multicolumn{2}{c|}{\textbf{\DSI}}
    & \multicolumn{2}{c|}{\textbf{\DSII}}
    & \multicolumn{2}{c|}{\textbf{\DSIII}}
    & \multicolumn{2}{c|}{\textbf{\DSIV}}
    & \multicolumn{2}{c|}{\textbf{\DSV}}
    & \multicolumn{2}{c|}{\textbf{\DSVI}}
    & \multicolumn{2}{c|}{\textbf{\DSVII}}
    & \multicolumn{2}{c|}{\textbf{\DSVIII}}
    & \multicolumn{2}{c}{\textbf{Average}} \\
    \textbf{System} 
        & \metricheadcov & \metricheadcov & \metricheadcov & \metricheadcov 
        & \metricheadcov & \metricheadcov & \metricheadcov & \metricheadcov
        & \metricheadcov \\
    \hline
    \rowcolor{gray!15}
    \multicolumn{19}{c}{\textbf{\textsl{LLM (GPT-4.1)}}} \\
    \hline
    Vanilla  & 23.39 & 83.82
         & 31.58 & 90.37
         & 16.05 & 82.56
         & 19.23 & 85.52
         & 23.29 & 83.83
         & 20.22 & 82.27
         & 19.33 & 85.32
         & 19.96 & 86.05
         &\cellcolor{red!24} 21.63 & 84.97 \\
    COT      & 21.68 & 84.32
         & 29.62 & 89.41
         & 17.23 & 84.37
         & 19.20 & 85.46
         & 23.64 & 84.35
         & 21.11 & 85.71
         & 19.24 & 86.14
         & 19.45 & 85.11
         & 21.40 & 85.61 \\

SC       & 21.96 & 84.58
         & 29.87 & 90.57
         & 16.07 & 84.97
         & 18.71 & 84.83
         & 23.97 & 85.03
         & 19.51 & 84.77
         & 19.00 & 85.99
         & 19.13 & 86.16
         & 21.03 & 85.86 \\

     Cite     & 21.58 & 84.66
     & 27.49 & 90.11
     & 14.94 & 76.96
     & 18.66 & 84.05
     & 23.55 & 84.13
     & 18.51 & 83.86
     & 17.01 & 84.09
     & 19.43 & 86.39
     & 20.15 & 84.28 \\

Deco     & 19.96 &84.96
         & 25.27 &86.02
         & 12.18 &81.69
         & 17.16 &85.26
         & 19.50 &79.80
         & 17.19 &81.68
         & 17.80 &85.83
         & 18.84 &85.78
         & 18.49 &83.88  \\

    QAG  & 20.91 & 84.09
         & 27.73 & 89.89
         & 14.59 & 84.95
         & 16.25 & 84.39
         & 22.41 & 84.30
         & 20.52 & 84.63
         & 18.31 & 85.70
         & 18.64 & 85.03
         & 19.92 & 85.37 \\

Plan     & 19.89 & 85.75
         & 23.71 & 89.34
         & 12.97 & 83.99
         & 15.42 & 84.59
         & 19.79 & 83.15
         & 18.41 & 85.15
         & 15.71 & 85.02
         & 18.43 & 85.28
         & 18.04 & 85.28 \\

    IR       & 22.80 & 86.27
         & 29.97 & 90.47
         & 17.56 & 85.58
         & 18.40 & 85.20
         & 23.93 & 85.03
         & 20.72 & 85.72
         & 19.32 & 86.11
         & 19.94 & 86.34
         & \cellcolor{red!16}21.58 & \cellcolor{red!24}86.34 \\

    E2A      & 21.66 & 85.19
         & 28.44 & 89.33
         & 18.72 & 85.02
         & 18.32 & 85.40
         & 22.90 & 84.83
         & 18.93 & 85.50
         & 18.34 & 85.92
         & 19.60 & 86.27
         & 20.87 & \cellcolor{red!16}85.93 \\

    \hline
    \rowcolor{gray!15}
    \multicolumn{19}{c}{\textbf{\textsl{Large Reasoning Models (LRM)}}} \\
    \hline
    o1       & 21.54 & 83.56
         & 25.50 & 89.02
         & 15.84 & 83.25
         & 17.38 & 83.52
         & 20.88 & 84.14
         & 17.67 & 83.53
         & 16.85 & 84.66
         & 17.66 & 85.50
         & 19.17 &84.65  \\

    o3       & 18.28 & 82.47
         & 22.67 & 86.84
         & 12.55 & 74.04
         & 16.31 & 84.61
         & 16.30 & 81.44
         & 18.44 & 84.69
         & 17.36 & 82.51
         & 15.51 & 84.28
         & 17.18 & 82.61 \\

    GPT-5    & 19.80 & 82.41
         & 25.58 & 87.67
         & 14.49 & 79.96
         & 16.41 & 84.06
         & 17.22 & 82.11
         & 19.63 & 84.64
         & 18.23 & 84.05
         & 15.86 & 83.83
         & 18.40 & 83.59 \\

    \hline
\end{tabular}}
\caption{2-shot performance across 8 datasets with ROUGE (average of ROUGE 1/2/L) and BERTScore (BS). The last two columns show per-system averages across all 8 datasets.}
\label{tab:coverage_results_8ds_2shot}
\end{table*}

\newcommand{\metricheadfaith}{\textbf{SumC} & \textbf{AS}}

\begin{table*}[ht]
\setlength{\tabcolsep}{3pt} 
\renewcommand{\arraystretch}{1.2}
\centering
\scriptsize 

\resizebox{\textwidth}{!}{
\begin{tabular}{l|cc|cc|cc|cc|cc|cc|cc|cc|cc}
    \hline
    & \multicolumn{2}{c|}{\textbf{\DSI}}
    & \multicolumn{2}{c|}{\textbf{\DSII}}
    & \multicolumn{2}{c|}{\textbf{\DSIII}}
    & \multicolumn{2}{c|}{\textbf{\DSIV}}
    & \multicolumn{2}{c|}{\textbf{\DSV}}
    & \multicolumn{2}{c|}{\textbf{\DSVI}}
    & \multicolumn{2}{c|}{\textbf{\DSVII}}
    & \multicolumn{2}{c|}{\textbf{\DSVIII}}
    & \multicolumn{2}{c}{\textbf{Avg}} \\
    \textbf{System}
        & \metricheadfaith & \metricheadfaith & \metricheadfaith & \metricheadfaith
        & \metricheadfaith & \metricheadfaith & \metricheadfaith & \metricheadfaith
        & \metricheadfaith \\
    \hline
    \rowcolor{gray!15}
    \multicolumn{19}{c}{\textbf{\textsl{LLM (GPT-4.1)}}} \\
    \hline
    Vanilla  & 45.49 & 83.49
         & 41.08 & 72.44
         & 33.77 & 45.56
         & 38.40 & 40.70
         & 25.24 & 56.25
         & 26.84 & 86.84
         & 27.35 & 98.38
         & 34.83 & 93.78
         & \cellcolor{red!16} 34.13 & \cellcolor{red!16} 72.18 \\

    COT      & 30.94 & 42.67 
         & 27.41 & 86.62 
         & 27.50 & 49.92 
         & 34.06 & 94.17 
         & 40.54 &  7.46 
         & 37.07 & 39.78 
         & 38.31 & 67.03 
         & 28.22 & 99.45 
 & 33.01 & 60.89 \\

    SC       & 40.44 &  6.39
         & 38.65 & 67.06
         & 28.47 & 40.43
         & 35.72 & 39.73
         & 24.25 & 53.12
         & 23.64 & 88.77
         & 23.30 & 95.93
         & 33.19 & 88.79
         & 30.96 & 60.03 \\

    Cite     & 43.91 & 11.92
         & 39.36 & 67.54
         & 31.75 & 44.63
         & 38.37 & 42.67
         & 31.97 & 60.19
         & 25.25 & 87.63
         & 24.81 & 96.99
         & 33.10 & 86.38
         & 33.57 & 62.24 \\

    Deco     & 43.67 & 66.06
         & 37.50 & 73.20
         & 26.84 & 37.86
         & 36.22 & 44.27
         & 24.61 & 55.42
         & 23.35 & 88.03
         & 23.94 & 98.64
         & 29.31 & 76.49
         & 30.68 & 67.25 \\

    QAG      & 31.59 & 40.70
         & 24.43 & 88.32
         & 26.96 & 56.02
         & 32.36 & 76.86
         & 40.90 & 29.66
         & 36.44 & 33.61
         & 39.37 & 72.69
         & 24.84 & 98.63
         & 32.11 & 62.06 \\

    Plan     & 38.91 & 10.86
         & 38.83 & 72.78
         & 25.98 & 32.38
         & 32.61 & 40.02
         & 23.85 & 56.64
         & 23.40 & 87.72
         & 22.21 & 97.83
         & 30.28 & 83.90
         & 29.51 & 60.27 \\

    IR        & 40.99 & 8.53 
          & 38.66 & 69.23 
          & 29.45 & 40.04 
          & 34.90 & 37.58 
          & 25.17 & 54.72 
          & 24.61 & 89.57 
          & 26.77 & 98.30 
          & 32.63 & 85.05 
          & 31.65 & 60.13 \\

    E2A      & 33.58 & 43.24 
         & 25.96 & 90.14 
         & 26.51 & 53.41 
         & 36.87 & 88.49 
         & 45.74 & 12.12 
         & 37.56 & 42.19 
         & 39.84 & 68.04 
         & 24.59 & 97.13 
         & 33.83 & 61.72 \\

    \hline
    \rowcolor{gray!15}
    \multicolumn{19}{c}{\textbf{\textsl{Large Reasoning Models (LRM)}}} \\
    \hline
    o1       & 31.24 & 39.83 
         & 27.61 & 86.52 
         & 26.02 & 54.86 
         & 34.67 & 94.46 
         & 44.23 & 31.71 
         & 38.02 & 51.92 
         & 38.82 & 68.96 
         & 29.09 & 97.30 
         & 33.71 & 65.70 \\

    o3   & 52.85 & 88.31
          & 37.98 & 69.22
          & 27.94 & 30.34
          & 34.62 & 47.07
          & 25.05 & 48.63
          & 23.29 & 88.07
          & 25.60 & 98.46
          & 35.34 & 86.93
          & 32.83 & \cellcolor{red!08} 69.63 \\

    GPT-5  & 48.59 & 91.57
           & 39.60 & 74.82
           & 30.56 & 35.32
           & 37.14 & 42.34
           & 27.38 & 56.30
           & 27.47 & 90.40
           & 27.32 & 99.37
           & 37.91 & 92.73
           &\cellcolor{red!08}  34.50 & \cellcolor{red!24} 72.86 \\

    \hline
\end{tabular}}
\caption{Faithfulness across 8 datasets with SummaC (SumC) and AlignScore (AS). The last two columns show per-system averages across the 8 datasets.}
\label{tab:faithfulness_results_8ds_sumc_as}
\end{table*}

\begin{table*}[ht]
\setlength{\tabcolsep}{3pt} 
\renewcommand{\arraystretch}{1.2}
\centering
\scriptsize 

\resizebox{\textwidth}{!}{
\begin{tabular}{l|cc|cc|cc|cc|cc|cc|cc|cc|cc}
    \hline
    & \multicolumn{2}{c|}{\textbf{\DSI}}
    & \multicolumn{2}{c|}{\textbf{\DSII}}
    & \multicolumn{2}{c|}{\textbf{\DSIII}}
    & \multicolumn{2}{c|}{\textbf{\DSIV}}
    & \multicolumn{2}{c|}{\textbf{\DSV}}
    & \multicolumn{2}{c|}{\textbf{\DSVI}}
    & \multicolumn{2}{c|}{\textbf{\DSVII}}
    & \multicolumn{2}{c|}{\textbf{\DSVIII}}
    & \multicolumn{2}{c}{\textbf{Avg}} \\
    \textbf{System}
        & \metricheadfaith & \metricheadfaith & \metricheadfaith & \metricheadfaith
        & \metricheadfaith & \metricheadfaith & \metricheadfaith & \metricheadfaith
        & \metricheadfaith \\
    \hline
    \rowcolor{gray!15}
    \multicolumn{19}{c}{\textbf{\textsl{LLM (GPT-4.1)}}} \\
    \hline
    Vanilla  & 35.56 & 47.12
         & 29.03 & 85.05
         & 32.21 & 29.35
         & 38.43 & 89.11
         & 43.36 & 26.07
         & 36.96 & 18.87
         & 39.04 & 66.56
         & 27.59 & 98.50
         &\cellcolor{red!16} 35.27 & 57.58 \\

    COT      & 30.94 & 42.67 
         & 27.41 & 86.62 
         & 27.50 & 49.92 
         & 34.06 & 94.17 
         & 40.54 &  7.46 
         & 37.07 & 39.78 
         & 38.31 & 67.03 
         & 28.22 & 99.45 
         & 33.01 & 60.89 \\

    SC       & 29.43 & 40.67
         & 25.14 & 86.47
         & 25.33 & 50.80
         & 32.52 & 75.70
         & 40.68 &  8.00
         & 35.39 & 40.53
         & 38.42 & 65.17
         & 24.61 & 98.44
         & 31.44 & 58.22 \\

    Cite     & 31.51 & 38.81
     & 24.68 & 86.83
     & 24.17 & 50.10
     & 34.73 & 92.01
     & 41.79 &  5.99
     & 36.05 & 38.20
     & 38.81 & 67.53
     & 24.66 & 97.84
     & 32.05 & 59.66 \\

    Deco     & 26.90 & 38.20
         & 24.36 & 88.02
         & 24.26 & 56.46
         & 30.96 & 87.23
         & 40.37 & 17.88
         & 33.68 & 37.27
         & 37.42 & 71.98
         & 25.04 & 98.59
         & 30.37 & 61.95 \\

    QAG     & 32.00 & 37.25
         & 25.34 & 86.88
         & 27.10 & 55.43
         & 34.32 & 85.87
         & 40.37 & 26.94
         & 35.86 & 40.98
         & 34.95 & 69.07
         & 25.62 & 98.91
         & 31.95 & 62.67 \\

        Plan     & 26.72 & 35.59
         & 24.30 & 87.17
         & 24.31 & 55.33
         & 30.97 & 86.16
         & 39.09 &  9.52
         & 33.01 & 41.17
         & 36.98 & 70.21
         & 23.46 & 99.45
         & 29.85 & 60.57 \\

        IR       & 29.62 & 40.42
         & 25.11 & 88.71
         & 24.59 & 44.35
         & 32.76 & 91.91
         & 39.87 &  5.65
         & 34.79 & 40.57
         & 37.57 & 65.76
         & 24.47 & 98.70
         & 31.10 & 59.51 \\

    E2A      & 32.85 & 37.28
         & 26.56 & 87.62
         & 26.53 & 42.49
         & 34.47 & 92.83
         & 42.96 &  8.68
         & 37.31 & 43.38
         & 37.77 & 65.07
         & 26.46 & 98.71
         & 33.11 & 59.51 \\

    \hline
    \rowcolor{gray!15}
    \multicolumn{19}{c}{\textbf{\textsl{Large Reasoning Models (LRM)}}} \\
    \hline
    o1       & 32.72 & 51.27 
         & 27.41 & 86.62 
         & 27.50 & 49.92 
         & 34.06 & 94.17 
         & 44.24 & 27.61 
         & 37.08 & 49.02 
         & 38.01 & 66.22 
         & 26.46 & 97.95 
        & 33.44 & \cellcolor{red!16}65.35 \\

    o3       & 27.72 & 42.15
         & 25.59 & 85.87
         & 26.54 & 35.54
         & 35.10 & 56.23
         & 53.58 & 85.22
         & 34.12 & 37.48
         & 39.65 & 66.61
         & 23.84 & 98.25
         & 33.27 & 63.42 \\

    GPT-5    & 32.69 & 62.82
         & 29.58 & 87.44
         & 34.39 & 21.41
         & 39.06 & 74.44
         & 45.90 & 85.55
         & 38.50 & 50.65
         & 40.71 & 78.62
         & 26.53 & 99.39
         &\cellcolor{red!24}35.92 &\cellcolor{red!24} 70.04 \\

    \hline
\end{tabular}}
\caption{2-shot Faithfulness across 8 datasets with SummaC (SumC) and AlignScore (AS). The last two columns show per-system averages across the 8 datasets.}
\label{tab:faithfulness_results_8ds_sumc_as_2shot}
\end{table*}

\section{Detailed Human Evaluation Results}

Here, we list the detailed Human Evaluation Scores (Completeness, Conciseness, Faithfulness)
of selected systems on three datasets. Table~\ref{tab:human_eval_compact_red} reports
the full human-annotated comparisons.
\begin{table*}[htbp]
\setlength{\tabcolsep}{2.5pt} 
\renewcommand{\arraystretch}{1.25}
\scriptsize 
\centering

\begin{tabular}{l|cccc|cccc|cccc|cccc}
\hline
& \multicolumn{4}{c|}{\textbf{arXiv}}
& \multicolumn{4}{c|}{\textbf{SciGen}}
& \multicolumn{4}{c|}{\textbf{WikiHow}}
& \multicolumn{4}{c}{\textbf{Average}} \\
\textbf{System}
& Con & Com & Fai & Avg
& Con & Com & Fai & Avg
& Con & Com & Fai & Avg
& Con & Com & Fai & Avg \\
\hline

Vanilla
& 3.75 & 4.05 & 4.15 & 3.98
& 4.20 & 3.60 & 3.90 & 3.90
& 4.15 & 3.80 & \cellcolor{red!24}4.60 & 4.18
& 4.03 & 3.82 & 4.22 & 4.02 \\

SC
& 4.05 & 3.75 & 4.20 & \cellcolor{red!24}4.00
& \cellcolor{red!24}4.30 & 3.55 & 4.10 & 3.98
& 4.25 & 3.85 & 4.50 & \cellcolor{red!24}4.20
& 4.20 & 3.72 & 4.27 & \cellcolor{red!24}4.06 \\

Deco
& 3.40 & 4.05 & 4.20 & 3.88
& 3.95 & 3.40 & \cellcolor{red!24}4.30 & 3.88
& 3.95 & 4.05 & 4.15 & 4.05
& 3.77 & 3.83 & 4.22 & 3.94 \\

E2A
& \cellcolor{red!24}4.50 & 2.60 & 3.70 & 3.60
& \cellcolor{red!24}4.30 & 2.60 & 4.00 & 3.63
& \cellcolor{red!24}4.45 & 3.10 & 4.25 & 3.93
& \cellcolor{red!24}4.42 & 2.77 & 3.98 & 3.72 \\

GPT-5
& 3.00 & \cellcolor{red!24}4.50 & \cellcolor{red!24}4.50 & \cellcolor{red!24}4.00
& 3.20 & \cellcolor{red!24}4.50 &\cellcolor{red!24} 4.30 & \cellcolor{red!24}4.00
& 3.30 & \cellcolor{red!24}4.45 & 4.45 & 4.07
& 3.17 & \cellcolor{red!24}4.48 & \cellcolor{red!24}4.42 & 4.02 \\

\hline
\end{tabular}

\caption{Human evaluation results across datasets (arXiv, SciGen, WikiHow). Highest values per column highlighted in red.}
\label{tab:human_eval_compact_red}
\end{table*}

\section{Detailed GEval Results}

Here, we list the detailed GEval Scores (Completeness, Conciseness, Faithfulness) for all
methods across all datasets. Table~\ref{tab:geval_dimensions} summarizes all GEval dimensions.

\newcommand{\metricheadgeval}{\textbf{Comp} & \textbf{Conc} & \textbf{Fact}}
\newcommand{\metricheadgevalavg}{\textbf{Comp} & \textbf{Conc} & \textbf{Fact} & \textbf{Avg}}

\begin{table*}[htbp]
\setlength{\tabcolsep}{3pt}
\renewcommand{\arraystretch}{1.2}
\centering
\scriptsize 

\begin{subtable}{\textwidth}
\centering
\caption{GEval-based Dimensions (1/2): \DSI, \DSII, \DSIII, \DSIV}
\begin{tabular}{l|ccc|ccc|ccc|ccc}
    \hline
    & \multicolumn{3}{c|}{\textbf{\DSI}}
    & \multicolumn{3}{c|}{\textbf{\DSII}}
    & \multicolumn{3}{c|}{\textbf{\DSIII}}
    & \multicolumn{3}{c}{\textbf{\DSIV}} \\
    \textbf{System} & \metricheadgeval & \metricheadgeval & \metricheadgeval & \metricheadgeval \\
    \hline
    \rowcolor{gray!15}
    \multicolumn{13}{c}{\textbf{\textsl{LLM (GPT-4.1)}}} \\
    \hline
    Vanilla & 4.82 & 4.60 & 4.97
        & 4.96 & 4.69 & 4.99
        & 4.92 & 4.69 & 4.95
        & 4.94 & 4.64 & 4.98 \\

    COT      & 4.90 & 4.23 & 4.98
        & 4.91 & 4.31 & 4.97
        & 4.87 & 4.28 & 4.94
        & 4.98 & 4.24 & 4.98 \\

    SC      & 4.89 & 4.63 & 5.00
        & 4.92 & 4.73 & 5.00
        & 4.91 & 4.83 & 4.99
        & 4.96 & 4.70 & 4.99 \\

    Cite    & 4.77 & 4.89 & 5.00
        & 4.93 & 4.44 & 4.98
        & 4.55 & 4.69 & 5.00
        & 4.92 & 4.94 & 4.99 \\

    Deco    & 4.98 & 4.21 & 5.00
        & 4.94 & 3.85 & 4.98
        & 4.90 & 3.91 & 4.94
        & 4.98 & 4.15 & 4.99 \\

    QA-G     & 5.00 & 4.07 & 5.00  
         & 4.96 & 3.77 & 5.00  
         & 4.82 & 4.07 & 4.99  
         & 4.98 & 3.90 & 4.99  \\
    Plan    & 4.88 & 4.14 & 5.00
        & 4.98 & 3.54 & 4.99
        & 4.80 & 4.00 & 4.97
        & 5.00 & 4.24 & 4.98 \\

    IR & 4.97 & 4.55 & 5.00
   & 4.96 & 4.46 & 5.00
   & 4.94 & 4.65 & 4.96
   & 4.98 & 4.51 & 4.98 \\

    E2A      & 4.89 & 4.88 & 5.00 & 4.96 & 4.85 & 5.00 & 4.66 & 4.88 & 5.00 & 4.86 & 4.78 & 4.99 \\
    \hline
    \rowcolor{gray!15}
    \multicolumn{13}{c}{\textbf{\textsl{Large Reasoning Models (LRM)}}} \\
    \hline
    o1      & 4.76 & 4.77 & 4.93
        & 4.82 & 4.61 & 4.94
        & 4.75 & 4.81 & 4.92
        & 4.91 & 4.88 & 4.96 \\

    o3      & 4.70 & 4.74 & 4.93
        & 4.88 & 4.81 & 4.96
        & 4.43 & 4.64 & 4.91
        & 4.99 & 4.88 & 4.99 \\

    GPT-5   & 4.79 & 4.47 & 5.00
        & 4.77 & 4.67 & 4.94
        & 4.55 & 4.52 & 4.88
        & 4.82 & 4.37 & 4.96 \\

    \hline
\end{tabular}
\end{subtable}

\begin{subtable}{\textwidth}
\centering
\caption{GEval-based Dimensions (2/2): \DSV, \DSVI, \DSVII, \DSVIII}
\begin{tabular}{l|ccc|ccc|ccc|ccc|cccc}
    \hline
    & \multicolumn{3}{c|}{\textbf{\DSV}}
    & \multicolumn{3}{c|}{\textbf{\DSVI}}
    & \multicolumn{3}{c|}{\textbf{\DSVII}}
    & \multicolumn{3}{c|}{\textbf{\DSVIII}}
    & \multicolumn{4}{c}{\textbf{Average}} \\
    \textbf{System} & \metricheadgeval & \metricheadgeval & \metricheadgeval & \metricheadgeval & \metricheadgevalavg \\
    \hline
    \rowcolor{gray!15}
    \multicolumn{17}{c}{\textbf{\textsl{LLM (GPT-4.1)}}} \\
    \hline
    Vanilla & 5.00 & 4.29 & 5.00
        & 4.94 & 4.56 & 4.99
        & 4.92 & 4.47 & 4.94
        & 4.97 & 4.61 & 4.97
        & 4.93 & 4.57 & 4.97 & 4.82 \\

    COT      & 5.00 & 4.05 & 5.00
        & 4.98 & 4.24 & 5.00
        & 5.00 & 3.17 & 4.98
        & 4.99 & 3.93 & 4.98
        & 4.95 & 4.06 & 4.98 & 4.66 \\

    SC      & 4.99 & 4.68 & 5.00
        & 4.95 & 4.59 & 4.98
        & 4.98 & 4.63 & 4.97
        & 4.84 & 4.76 & 4.99
        & 4.93 & 4.69 & \cellcolor{red!24}4.99 &\cellcolor{red!24} 4.87 \\

    Cite    & 4.81 & 4.86 & 4.98
        & 4.83 & 4.84 & 4.98
        & 4.93 & 4.88 & 4.97
        & 4.71 & 4.80 & 4.98
        & 4.81 & 4.79 & 4.98 &\cellcolor{red!16} 4.86 \\

    Deco    & 4.98 & 3.91 & 4.97
        & 4.92 & 4.10 & 4.98
        & 5.00 & 4.23 & 5.00
        & 4.86 & 4.20 & 4.96
        & 4.95 & 4.07 & 4.98 & 4.66 \\

    QA-G     & 3.83 & 3.56 & 4.97 
         & 1.93 & 3.97 & 4.98 
         & 2.03 & 2.50 & 4.85  
         & 4.38 & 3.92 & 4.99  
         & 3.99 & 3.72 & 4.97 & 4.23 \\
    Plan    & 4.98 & 4.38 & 5.00
        & 4.91 & 4.14 & 4.95
        & 4.95 & 4.04 & 4.99
        & 4.99 & 3.69 & 4.97
        & 4.93 & 4.02 & 4.98 & 4.65 \\

    IR & 5.00 & 4.69 & 5.00
   & 4.96 & 4.51 & 4.97
   & 5.00 & 4.53 & 5.00
   & 4.91 & 4.30 & 4.99
   & \cellcolor{red!24}4.97 & 4.53 &\cellcolor{red!24} 4.99 & 4.83 \\

    E2A & 4.67 & 4.92 & 4.99 & 4.68 & 4.82 & 4.99 & 4.24 & 4.83 & 4.95 & 4.47 & 4.87 & 4.99 & 4.68 &\cellcolor{red!24} 4.85 &\cellcolor{red!24} 4.99 & 4.84 \\

    \hline
    \rowcolor{gray!15}
    \multicolumn{17}{c}{\textbf{\textsl{Large Reasoning Models (LRM)}}} \\
    \hline
    o1      & 4.97 & 4.88 & 5.00
        & 4.80 & 4.63 & 4.92
        & 4.94 & 4.87 & 4.98
        & 4.88 & 4.62 & 4.92
        & 4.85 & 4.76 & 4.95 & \cellcolor{red!24}4.87 \\

    o3      & 5.00 & 4.87 & 5.00
        & 4.86 & 4.67 & 4.97
        & 4.97 & 4.73 & 5.00
        & 4.98 & 4.40 & 4.98
        & 4.85 & 4.72 & 4.97 & 4.84 \\

    GPT-5   & 5.00 & 4.33 & 5.00
        & 4.79 & 4.20 & 4.98
        & 4.93 & 4.25 & 4.99
        & 4.97 & 4.35 & 4.96
        & 4.83 & 4.40 & 4.96 & 4.73 \\

    \hline
\end{tabular}
\end{subtable}

\caption{Performance across 8 datasets with GEval dimensions: Completeness (Comp), Conciseness (Conc), and Factualness (Fact).}
\label{tab:geval_dimensions}
\end{table*}



\section{Compression Ratio and Abstractiveness}

We also provide the Compression Ratio and Abstractiveness of the summaries.
Table~\ref{tab:ca_results_8ds_oneblock} shows the complete results.
\newcommand{\pct}[1]{\fpeval{round(100*(#1),2)}\%}

\newcommand{\metricheadCA}{\textbf{CR} & \textbf{Abs.}}

\begin{table*}[ht]
\setlength{\tabcolsep}{3pt}
\renewcommand{\arraystretch}{1.2}
\centering
\scriptsize

\resizebox{\textwidth}{!}{
\begin{tabular}{l|cc|cc|cc|cc|cc|cc|cc|cc|cc}
    \hline
    & \multicolumn{2}{c|}{\textbf{\DSI}}
    & \multicolumn{2}{c|}{\textbf{\DSII}}
    & \multicolumn{2}{c|}{\textbf{\DSIII}}
    & \multicolumn{2}{c|}{\textbf{\DSIV}}
    & \multicolumn{2}{c|}{\textbf{\DSV}}
    & \multicolumn{2}{c|}{\textbf{\DSVI}}
    & \multicolumn{2}{c|}{\textbf{\DSVII}}
    & \multicolumn{2}{c|}{\textbf{\DSVIII}}
    & \multicolumn{2}{c}{\textbf{Average}} \\
    \textbf{System}
        & \metricheadCA & \metricheadCA & \metricheadCA & \metricheadCA
        & \metricheadCA & \metricheadCA & \metricheadCA & \metricheadCA
        & \metricheadCA \\
    \hline
    \rowcolor{gray!15}
    \multicolumn{19}{c}{\textbf{\textsl{LLM (GPT-4.1)}}} \\
    \hline
Vanilla  & \pct{0.210982} & \pct{0.528449}  
         & \pct{0.636495} & \pct{0.727023}  
         & \pct{0.256150} & \pct{0.678596}  
         & \pct{0.312170} & \pct{0.585883}  
         & \pct{0.065390} & \pct{0.519755}  
         & \pct{0.112905} & \pct{0.520803}  
         & \pct{0.086801} & \pct{0.679372}  
         & \pct{0.631989} & \pct{0.664571}  
         & \pct{0.28911025} & \pct{0.6130565} \\ 

COT      & \pct{0.265565} & \pct{0.546714}  
         & \pct{0.734108} & \pct{0.817190}  
         & \pct{0.287268} & \pct{0.731231}  
         & \pct{0.372665} & \pct{0.650212}  
         & \pct{0.072844} & \pct{0.527956}  
         & \pct{0.142158} & \pct{0.538573}  
         & \pct{0.147758} & \pct{0.740458}  
         & \pct{1.009655} & \pct{0.700211}  
         & \pct{0.379002625} & \pct{0.656568125} \\ 

SC       & \pct{0.207254} & \pct{0.546765}  
         & \pct{0.610380} & \pct{0.746996}  
         & \pct{0.234074} & \pct{0.651333}  
         & \pct{0.308797} & \pct{0.608489}  
         & \pct{0.046625} & \pct{0.482475}  
         & \pct{0.108336} & \pct{0.532416}  
         & \pct{0.083831} & \pct{0.683532}  
         & \pct{0.486398} & \pct{0.693777}  
         & \pct{0.260711875} & \pct{0.618222875} \\ 

Cite     & \pct{0.178480} & \pct{0.571084}  
         & \pct{0.728084} & \pct{0.793273}  
         & \pct{0.223176} & \pct{0.638343}  
         & \pct{0.261317} & \pct{0.646377}  
         & \pct{0.034530} & \pct{0.472967}  
         & \pct{0.090607} & \pct{0.541132}  
         & \pct{0.069142} & \pct{0.689062}  
         & \pct{0.508597} & \pct{0.717009}  
         & \pct{0.261741625} & \pct{0.633655875} \\ 

Deco     & \pct{0.241152} & \pct{0.611525}  
         & \pct{0.898784} & \pct{0.800223}  
         & \pct{0.337426} & \pct{0.721763}  
         & \pct{0.349323} & \pct{0.678221}  
         & \pct{0.074366} & \pct{0.505270}  
         & \pct{0.133433} & \pct{0.574422}  
         & \pct{0.091603} & \pct{0.706215}  
         & \pct{0.682423} & \pct{0.739800}  
         & \pct{0.35106375} & \pct{0.667179875} \\ 

QAG      & \pct{0.300344} & \pct{0.533402}  
         & \pct{0.831533} & \pct{0.766776}  
         & \pct{0.384186} & \pct{0.684530}  
         & \pct{0.428038} & \pct{0.590403}  
         & \pct{0.068909} & \pct{0.493277}  
         & \pct{0.164403} & \pct{0.517579}  
         & \pct{0.126153} & \pct{0.661940}  
         & \pct{1.021179} & \pct{0.659223}  
         & \pct{0.415593125} & \pct{0.61339125} \\ 

Plan     & \pct{0.245215} & \pct{0.602935}  
         & \pct{0.918963} & \pct{0.822878}  
         & \pct{0.344332} & \pct{0.676290}  
         & \pct{0.377154} & \pct{0.656643}  
         & \pct{0.050217} & \pct{0.556617}  
         & \pct{0.128399} & \pct{0.575505}  
         & \pct{0.093026} & \pct{0.723847}  
         & \pct{1.073159} & \pct{0.761927}  
         & \pct{0.403808125} & \pct{0.67208025} \\ 

IR       & \pct{0.233506} & \pct{0.516461}  
         & \pct{0.695402} & \pct{0.750685}  
         & \pct{0.284014} & \pct{0.650465}  
         & \pct{0.339572} & \pct{0.586099}  
         & \pct{0.047733} & \pct{0.470309}  
         & \pct{0.120266} & \pct{0.506187}  
         & \pct{0.089021} & \pct{0.667374}  
         & \pct{0.729223} & \pct{0.647109}  
         & \pct{0.317842125} & \pct{0.599836125} \\ 

E2A      & \pct{0.176590} & \pct{0.535485}  
         & \pct{0.609627} & \pct{0.751230}  
         & \pct{0.211762} & \pct{0.673627}  
         & \pct{0.239797} & \pct{0.594061}  
         & \pct{0.040472} & \pct{0.484405}  
         & \pct{0.095559} & \pct{0.526003}  
         & \pct{0.075594} & \pct{0.684695}  
         & \pct{0.428948} & \pct{0.667946}  
         & \pct{0.234793625} & \pct{0.6146815} \\ 

    \hline

    \rowcolor{gray!15}
    \multicolumn{19}{c}{\textbf{\textsl{Large Reasoning Models (LRM)}}} \\
    \hline

o1       & \pct{0.192028} & \pct{0.546204}  
         & \pct{0.677481} & \pct{0.753402}  
         & \pct{0.240786} & \pct{0.695552}  
         & \pct{0.288295} & \pct{0.607784}  
         & \pct{0.040307} & \pct{0.547837}  
         & \pct{0.105059} & \pct{0.530251}  
         & \pct{0.070915} & \pct{0.701585}  
         & \pct{0.679026} & \pct{0.658884}  
         & \pct{0.286737125} & \pct{0.630187375} \\ 

o3       & \pct{0.211785} & \pct{0.596299}  
         & \pct{0.633669} & \pct{0.768469}  
         & \pct{0.247826} & \pct{0.708040}  
         & \pct{0.321895} & \pct{0.661028}  
         & \pct{0.072747} & \pct{0.599614}  
         & \pct{0.122889} & \pct{0.559218}  
         & \pct{0.093660} & \pct{0.694928}  
         & \pct{0.883925} & \pct{0.651097}  
         & \pct{0.3235495} & \pct{0.654836625} \\ 

GPT-5    & \pct{0.257201} & \pct{0.523518}  
         & \pct{0.693616} & \pct{0.707981}  
         & \pct{0.271777} & \pct{0.648567}  
         & \pct{0.393022} & \pct{0.575884}  
         & \pct{0.106250} & \pct{0.563541}  
         & \pct{0.153547} & \pct{0.510826}  
         & \pct{0.121298} & \pct{0.659003}  
         & \pct{0.896624} & \pct{0.567692}  
         & \pct{0.361666875} & \pct{0.5946265} \\ 

    \hline
\end{tabular}}
\caption{Compression Ratio (CR) and Abstractiveness (Abs.) of the summaries generated by 0-shot experiments are reported across eight datasets, with all values presented as percentages rounded to two decimal places. The last two columns show the average results of each system over the eight datasets.}
\label{tab:ca_results_8ds_oneblock}
\end{table*}

\section{Reasoning Strategy Selection}
\label{app:recommended_methods}
Based on our comprehensive evaluation, we summarize the recommended reasoning strategies for each dataset and domain. Table~\ref{tab:recommended_methods_simplified} highlights the most effective method for maximizing either summary quality or factual faithfulness under different settings.

\begin{table*}[ht]
    \centering
    \footnotesize
    
    \begin{tabular}{l l c c c c}
        \toprule
        \multirow{2}{*}{\textbf{Dataset}} & \multirow{2}{*}{\textbf{Domain}} & \multicolumn{2}{c}{\textbf{0-shot Setting}} & \multicolumn{2}{c}{\textbf{2-shot Setting}} \\
        \cmidrule(lr){3-4} \cmidrule(lr){5-6}
         & & \textbf{Quality} & \textbf{Faithfulness} & \textbf{Quality} & \textbf{Faithfulness} \\
        \midrule
        
        \multicolumn{6}{c}{\textit{Short-Form Documents (SDS)}} \\
        \midrule
        CNN/DM & News & \textbf{Cite} & \textbf{GPT-5} & \textbf{IR} & \textbf{Vanilla} \\
        SAMSum & Dialogue & \textbf{SC} & \textbf{GPT-5} & \textbf{SC} & \textbf{GPT-5} \\
        Reddit & Social Media & \textbf{SC} & \textbf{Vanilla} & \textbf{IR} & \textbf{QAG} \\
        WikiHow & Knowledge Base & \textbf{Cite} & \textbf{E2A} & \textbf{Vanilla} & \textbf{CoT} \\
        \midrule
        
        \multicolumn{6}{c}{\textit{Long-Form Documents (MDS / LNS)}} \\
        \midrule
        ArXiv & Scientific Paper & \textbf{IR} & \textbf{Cite} & \textbf{SC} & \textbf{GPT-5} \\
        Multi-News & News & \textbf{IR} & \textbf{GPT-5} & \textbf{IR} & \textbf{GPT-5} \\
        BookSum & Novels & \textbf{IR} & \textbf{GPT-5} & \textbf{CoT} & \textbf{GPT-5} \\
        \midrule
        
        \multicolumn{6}{c}{\textit{Table-to-Text (TTS)}} \\
        \midrule
        SciGen & Scientific Paper & \textbf{SC} & \textbf{GPT-5} & \textbf{IR} & \textbf{CoT} \\
        
        \bottomrule
    \end{tabular}
    
    \caption{Recommended reasoning strategies across datasets and domains. Methods are selected based on the \textbf{average ranking} of constituent metrics: \textit{Quality} considers both ROUGE and BERTScore, while \textit{Faithfulness} considers SummaC and AlignScore. The best method (lowest average rank) is shown for each setting.}
    \label{tab:recommended_methods_simplified}
\end{table*}

\section{Case Study}
\label{app:case_study}

To study the potential hallucination caused by the reasoning~\cite{chen2026residual,zhang2026stable}, here we show the full example of the generated summaries in Table~\ref{tab:case_study} from the Wikihow dataset, together with the golden reference. 

\begin{table*}[ht!]
\footnotesize
\begin{tabular}{p{1.2cm}p{14.0cm}}
\toprule
\textbf{Document} & 
Clean the knife and cutting board with hot water and soap, and be sure to rinse both off thoroughly. Your vegetables don’t need to be washed with soap, but rinse with hot water and consider scrubbing with a cloth or brush. Use a sharp, non-serrated knife for cutting. A chopping knife works well but is not required. It’s important to only use a sharp knife so you can make clean cuts. Potatoes and carrots are often peeled before being julienned. It is up to you depending on how you plan to eat them. If you don’t peel your potatoes, at least make sure buds or eyes are removed from the skin. Use a peeler or a paring knife to remove the skin. Cut off the stem and bottoms when necessary. Slice off one side at a time so you are left with a rectangular chunk of vegetable. You can throw the scraps away, or you can slice them up and mix them in with the uniform pieces you’ll cut later. This is especially helpful for potatoes, zucchini, and squash. If you are going to be eating the pieces raw, they can be a little longer, but \textcolor{blue}{if you’re cooking with them, $2$--$3$ inches is best}. Each chunk does not have to be the exact same length, just divide the vegetable evenly. The general rule for juliennes is that you’d be able to eat them in one bite, but you can cut them into longer chunks if you want. When cutting juliennes, $\tfrac{1}{8}\,\text{in}$ $(0.3\,\text{cm})$ to $1$ be careful not to cut yourself as you slice. \textcolor{blue}{Flip the slices} you’ve just cut so that they are piled on top of each other. Then cut them into \textcolor{blue}{sticks that are $\tfrac{1}{8}\,\text{in}$ $(0.3\,\text{cm})$ to $\tfrac{1}{16}\,\text{in}$ $(0.15\,\text{cm})$ thick}, again. The goal with juliennes is to get them all to a fairly uniform size.
 \\
\midrule
\textbf{Reference} & 
Wash the knife, cutting board, and vegetables. Peel the vegetable if you want to. Cut rounded vegetables into a rectangle. chop the vegetable into \textcolor{blue}{chunks that are 2-3 inches ( 5-8 cm ) long}. Slice the vegetable down the length. \textcolor{blue}{stack 2-3 slices} and cut again .
 \\
\midrule
\textbf{Vanilla} & 
Clean the knife and cutting board with hot water and soap, rinsing thoroughly. Rinse vegetables with hot water, scrubbing if needed; do not use soap. Use a sharp, non-serrated knife, or optionally a chopping knife, for clean cuts. Peel potatoes and carrots if desired, and remove eyes from potatoes. Use a peeler or paring knife to remove skin and cut off stems and bottoms as needed. Slice vegetables into rectangular chunks, discarding or saving scraps. For julienne cuts, divide vegetables evenly; \textcolor{blue}{raw pieces can be longer, but cooked ones should be $2$--$3$ inches}. Cut slices into \textcolor{blue}{sticks $\tfrac{1}{8}\,\text{in}$ $(0.3\,\text{cm})$ to $\tfrac{1}{16}\,\text{in}$ $(0.15\,\text{cm})$ thick}, \textcolor{blue}{stacking slices} for uniformity, and always be careful not to cut yourself. Aim for uniform, bite-sized pieces.
 \\
\midrule
\textbf{SC} & 
To julienne vegetables, clean your knife and cutting board thoroughly with hot water and soap, and rinse well. Rinse vegetables with hot water and scrub if needed, but do not use soap. Use a sharp, non-serrated knife for clean cuts; peeling potatoes and carrots is optional, but remove any eyes or buds if not peeling. Cut off stems and bottoms as needed, slice vegetables into rectangular chunks, and either discard or use scraps. For raw pieces, longer cuts are acceptable; \textcolor{blue}{for cooking, $2$--$3$ inch lengths are best}. Aim for uniform, bite-sized \textcolor{blue}{sticks about $\tfrac{1}{8}$ to $\tfrac{1}{16}$ inch thick}, \textcolor{blue}{stacking slices} before cutting, and always be careful not to cut yourself.
 \\
\midrule
\textbf{Deco} & 
Begin by thoroughly cleaning your knife and cutting board with hot water and soap, and rinsing vegetables with hot water to ensure hygiene. Use a sharp, non-serrated knife for safe, clean cuts. Prepare vegetables by peeling them if desired, removing unwanted parts like stems, bottoms, or potato eyes, and cutting them into manageable rectangular chunks based on your intended use and preference. For julienning, \textcolor{blue}{stack the slices} and cut them into \textcolor{blue}{uniform sticks about $\tfrac{1}{8}$ to $\tfrac{1}{16}$ inch thick} \textcolor{red}{and $2$--$3$ inches long}, aiming for consistency in size to ensure even cooking and presentation.
 \\
\midrule
\textbf{E2A} & 
To julienne vegetables, first clean your knife and cutting board thoroughly. Use a sharp, non-serrated knife for precise cuts. Whether or not you peel vegetables depends on how you plan to eat them. Slice off one side at a time to create rectangular chunks, aiming for pieces that are generally bite-sized, though longer chunks are acceptable if preferred. \textcolor{red}{The key is to cut the juliennes to a fairly uniform size.}
 \\
\midrule
\textbf{GPT-5} & 
- Clean knife and cutting board with hot, soapy water; rinse thoroughly. Rinse vegetables with hot water (no soap); scrub with a cloth/brush if needed. \newline
- Use a sharp, non-serrated knife (a chopping knife works). Sharpness ensures clean cuts. \newline
- Peel potatoes and carrots if desired; always remove potato eyes. Use a peeler or paring knife. Trim stems and bottoms as needed. \newline
- Square the vegetable by slicing off sides to form a rectangular chunk; discard or use trimmings. This is especially helpful for potatoes, zucchini, and squash. \newline
- Cut the chunk into even slabs, \textcolor{blue}{stack them}, then slice into \textcolor{blue}{uniform sticks about $\tfrac{1}{8}$--$\tfrac{1}{16}$ inch ($\approx 0.3$--$0.15$ cm) thick}. Aim for one-bite, fairly uniform pieces; be careful not to cut yourself. \newline
- Length: \textcolor{blue}{for cooking, $2$--$3$ inches is best; raw pieces can be a little longer}. Divide the vegetable into evenly sized chunks.
 \\
\bottomrule
\end{tabular}
\caption{Case study comparing summaries generated by different reasoning strategies against a human reference. \textcolor{blue}{Blue text} in summaries shows their better faithfulness to the source, as they manage to keep these important procedural details that the reference missed.  \textcolor{red}{red text} marks hallucinations or omissions, such as \textbf{Deco}'s failure to recognize conditional branches (generalizing cooking length to all cases) or \textbf{E2A}'s complete omission of the final cutting step.}
\label{tab:case_study}
\end{table*}

\section{Human Evaluation Guidelines}
\label{Appendix:Human Evaluation}
Our human evaluation used two PhD students, each independently rating summaries using the rubric in Table~\ref{tab:human_eval_guidelines} on a 1–5 scale for Conciseness, Completeness, and Faithfulness. We report the average of their scores across examples. 
Here we list the full human evaluation guidelines used for our manual annotation study,
including rating scales and definitions.

\begin{table*}[ht]
\centering
\small
\renewcommand{\arraystretch}{1.3}
\begin{tabular}{p{2.8cm} p{1cm} p{10cm}}
\hline
\textbf{Criterion} & \textbf{Score} & \textbf{Description} \\
\hline
\multirow{5}{*}{\textbf{Conciseness}} 
& 5 & The summary is highly concise, containing only essential information with no redundancy or irrelevant content. \\
& 4 & Mostly concise with minor redundancy; still clear and compact overall. \\
& 3 & Contains some unnecessary details or minor repetition that slightly affects brevity. \\
& 2 & Wordy or includes multiple redundant or off-topic parts; could be shortened significantly. \\
& 1 & Very verbose or includes large amounts of irrelevant content; lacks focus and compactness. \\
\hline
\multirow{5}{*}{\textbf{Completeness}} 
& 5 & Fully covers all key points and information in the source document; no important content omitted. \\
& 4 & Covers most key information, with only minor omissions that do not affect overall understanding. \\
& 3 & Misses some moderately important points; the reader may lose some aspects of the original meaning. \\
& 2 & Covers only a portion of the important content; several key ideas missing. \\
& 1 & Severely incomplete; fails to convey most of the main ideas of the source. \\
\hline
\multirow{5}{*}{\textbf{Faithfulness}} 
& 5 & Entirely faithful to the source; no hallucination, distortion, or factual errors. \\
& 4 & Mostly faithful, with only minor wording deviations or small factual imprecision. \\
& 3 & Some inaccuracies or inferred details not explicitly supported by the source. \\
& 2 & Several incorrect or misleading statements that distort the original meaning. \\
& 1 & Largely unfaithful; contains major fabrications or contradictions with the source content. \\
\hline
\end{tabular}
\caption{Human evaluation guidelines for summarization quality. Scores range from 1 to 5, with 0.5-point increments allowed for finer judgment.}
\label{tab:human_eval_guidelines}
\end{table*}

\section{Emerging Reasoning Directions for Future Adaptation}
\label{app:future_reasoning}
Beyond the methods benchmarked in this work, several recent directions may be adapted to abstractive summarization in future studies. Integrating Retrieval-Augmented Generation (RAG)~\cite{compselect} with reasoning-enhanced summarization represents a promising future direction, wherein relevant external passages can be dynamically retrieved to supplement key content segments with additional factual detail~\cite{zhang2024adacomp}. Moreover, recent analyses of logical phase transitions indicate that reasoning behavior can degrade abruptly under certain regimes, which is relevant to stability-aware summarization design~\cite{song10}.

At the reasoning-policy level, neuron-level mixture routing has been explored as a way to make reasoning trajectories more explainable and controllable~\cite{dong2026neureasonerexplainablecontrollableunified}, while error-driven refinement strategies provide another perspective on improving first-pass solution quality~\cite{jiang2026foeforesterrorsmakes}. Reasoning-path search approaches are also promising for selecting stronger inference traces before generation~\cite{ling2026neuralchainofthoughtsearchsearching, chen2026actormindemulatinghumanactor}. Semantic-aware frameworks for logical reasoning may further help control intermediate reasoning states during summary construction~\cite{song8}.

We also note that progress in parameter-efficient adaptation may offer practical implementation benefits when transferring new reasoning modules to summarization settings~\cite{dong2025aurora}. Meanwhile, advances in robust cross-modal retrieval and alignment can provide transferable architectural ideas, even when the source tasks are not summarization-specific~\cite{ReTrack,HABIT}. More broadly, work on long-horizon multi-agent dynamics suggests another possible route for modeling planner-decoder interactions in iterative summarization pipelines~\cite{song12}.

\section{Prompts}
\label{sec:prompt}
To ensure ethical transparency and reproducibility, we disclose the prompts we used to deploy each
reasoning strategy. The full templates are shown in Figures~\ref{fig:prompt_vanilla},
\ref{fig:prompt_cot_news}, \ref{fig:prompt_sc_full}, \ref{fig:prompt_cite},
\ref{fig:prompt_deco}, \ref{fig:prompt_qag}, \ref{fig:prompt_plan},
\ref{fig:prompt_ir}, and \ref{fig:prompt_e2a}.
\clearpage
\onecolumn

\captionsetup[figure]{justification=justified,singlelinecheck=false}
\tcbset{
  breakable,
  colback=gray!3,
  colframe=black!35,
  boxrule=0.4pt,
  arc=1mm,
  left=1mm,
  right=1mm,
  top=0.8mm,
  bottom=0.8mm
}
\sloppy

\subsection*{Vanilla and CoT}

\begin{tcolorbox}[title=Vanilla Prompt]
\small
\textbf{Instruction:} \\
You are a skilled analyst tasked with producing a concise and accurate summary from the given text.
The summary must preserve all essential facts, maintain logical flow, and avoid introducing information not present in the text.

\vspace{0.5em}
\textbf{Formatting Constraint:} \\
Then write ONLY the final summary text (no lists, no bullets, no metadata).
Aim to be as concise as possible while fully covering the key points.

\vspace{0.5em}
\textbf{[Begin Document]} \\
\{\textit{Input Document Content Here}\} \\
\textbf{[End Document]}

\vspace{1em}
 
    SUMMARY:
\end{tcolorbox}
\noindent\begin{minipage}{\textwidth}
\captionof{figure}{The Vanilla prompt used for the baseline setting. It provides general summarization constraints without specific reasoning instructions.}\label{fig:prompt_vanilla}
\end{minipage}

\vspace{2em} 

\begin{tcolorbox}[title=Chain-of-Thought (CoT) Prompt (News Domain)]
\small
\textbf{Instruction:} \\
You are a skilled analyst tasked with producing a concise and accurate summary from the given text.
The summary must preserve all essential facts, maintain logical flow, and avoid introducing information not present in the text.

\vspace{0.5em}
\textbf{Reasoning Guidance (Injected by Domain):} \\
First, internally (do NOT output lists), identify the following to guide your reasoning:
\begin{itemize}
    \setlength\itemsep{0em}
    \item important entities (people, organizations, places)
    \item important dates or time expressions
    \item key events or actions
    \item the outcome or conclusion of these events
\end{itemize}
\textit{Writing guidance: Keep chronology clear; avoid speculation; reflect the article's scope and tone.}

\vspace{0.5em}
\textbf{Formatting Constraint:} \\
Then write ONLY the final summary text (no lists, no bullets, no metadata).
Aim to be as concise as possible while preserving essentials.

\vspace{0.5em}
\textbf{[Begin Document]} \\
\{\textit{Input Document Content Here}\} \\
\textbf{[End Document]} 

\vspace{1em}
 
    SUMMARY:
\end{tcolorbox}
\noindent\begin{minipage}{\textwidth}
\captionof{figure}{The Chain-of-Thought (CoT) prompt template adapted for the \textbf{News} domain. The model identifies key entities and timelines internally before generating the summary.}\label{fig:prompt_cot_news}
\end{minipage}

\subsection*{Self-Consistency (SC)}
    \begin{tcolorbox}[title=Self-Consistency (SC) Step 1: Candidate Generation]
    \small
    \textbf{Instruction:} \\
You are a skilled analyst tasked with producing a concise and accurate summary from the given text.
The summary must preserve all essential facts, maintain logical flow, and avoid introducing information not present in the text.

\vspace{0.5em}
\textbf{Formatting Constraint:} \\
Then write ONLY the final summary text (no lists, no bullets, no metadata).
Aim to be as concise as possible while fully covering the key points.

\vspace{0.5em}
\textbf{[Begin Document]} \\
\{\textit{Input Document Content Here}\} \\
\textbf{[End Document]}

\vspace{1em}
 
    SUMMARY:
    \end{tcolorbox}

    \vspace{1em} 

    \begin{tcolorbox}[title=Self-Consistency (SC) Step 2: Evaluation and Selection]
    \small
    \textbf{Instruction:} \\
    You are a rigorous summarization evaluator. Given a DOCUMENT and three candidate abstractive summaries (A/B/C), score each summary and select ONE best candidate.
    
    \vspace{0.5em}
    \textbf{Evaluation Rubric:} \\
    For each candidate, assign sub-scores based on the following weights:
    \begin{itemize}
        \setlength\itemsep{0em}
        \item \textbf{Faithfulness (0.35)}: no hallucination; consistent with DOCUMENT.
        \item \textbf{Coverage (0.35)}: captures all core ideas; no major omissions.
        \item \textbf{Coherence (0.15)} / \textbf{Concision (0.15)}.
    \end{itemize}
    
    \vspace{0.5em}
    \textbf{Output Constraint:} \\
    Return ONLY a JSON object with the following keys:
    \begin{itemize}
        \setlength\itemsep{0em}
        \item \texttt{scores}: object with keys A/B/C.
        \item \texttt{winner}: one of 'A', 'B', 'C'.
        \item \texttt{reason}: explanation ($\le$ 50 words).
        \item \texttt{final\_summary}: EXACT text of the winning candidate.
    \end{itemize}
    
    \vspace{0.5em}
    \textbf{Input Context:} \\
    DOCUMENT: \{...\} \quad CANDIDATE\_A/B/C: \{...\}
    \end{tcolorbox}
    \captionof{figure}{Prompts used for the Self-Consistency (SC) method. \textbf{Top:} The prompt for generating candidate summaries (Step 1). \textbf{Bottom:} The prompt for the evaluator to score and select the best summary (Step 2).}\label{fig:prompt_sc_full}

\subsection*{Cited Summarization (Cite)}
\begin{tcolorbox}[title=Cited Summarization (Cite) Prompt]
\small
\textbf{Instruction:} \\
You are a faithful abstractive summarizer and evidence aligner.
Write an abstractive summary for the DOCUMENT and, for each summary sentence, cite up to 3 supporting sentences from the provided SENTENCES array and explain briefly why it is important.

\vspace{0.5em}
\textbf{Guidelines:} \\
\begin{itemize}
    \setlength\itemsep{0em}
    \item \textbf{Faithfulness}: Do not invent content not supported by DOCUMENT.
    \item \textbf{Citations}: Every summary sentence must have 1--3 supports from the SENTENCES list.
    \item \textbf{Paraphrase}: Do not copy support sentences verbatim (short quotes allowed).
    \item \textbf{Concision}: Use the minimum length that still covers key ideas.
\end{itemize}

\vspace{0.5em}
\textbf{Output Constraint (JSON):} \\
Output ONLY a JSON object matching the schema:
\begin{itemize}
    \setlength\itemsep{0em}
    \item \texttt{summary\_text}: string (the final abstractive summary).
    \item \texttt{alignments}: array of objects linking summary parts to evidence:
    \begin{itemize}
        \item \texttt{summary\_id}: integer (matching the summary sentence).
        \item \texttt{support}: array of indices from the SENTENCES list.
        \item \texttt{importance\_reason}: string (explanation).
        \item \texttt{support\_strength}: integer [1, 5].
    \end{itemize}
\end{itemize}

\vspace{0.5em}
\textbf{Input Context:} \\
SENTENCES (0-based array): ["Sentence 1...", "Sentence 2...", ...] \\
DOCUMENT: \{\textit{Input Document Content Here}\}
\end{tcolorbox}
\captionof{figure}{The prompt for Cited Summarization (Cite). It requires the model to generate the summary and simultaneously provide evidence alignments (indices from the source sentences) in a structured JSON format.}\label{fig:prompt_cite}

\subsection*{Decomposition (Deco)}
    \begin{tcolorbox}[title=Decomposition (Deco) Step 1: Chunking \& Local Summarization]
    \small
    \textbf{Instruction:} \\
    You are a planner and summarizer. Decompose the document into coherent contiguous chunks, then write a concise abstractive summary and a set of short supporting contexts for each chunk.
    
    \vspace{0.5em}
    \textbf{Task Guidelines:} \\
    \begin{itemize}
        \setlength\itemsep{0em}
        \item \textbf{Decompose}: Split the document into $M$ contiguous chunks (adaptively, typically 5--20 sentences each).
        \item \textbf{Local Summary}: For each chunk, write an abstractive summary.
        \item \textbf{Contexts}: Provide 3--6 short supporting contexts (key facts needed for coherence) for each chunk.
    \end{itemize}
    
    \vspace{0.5em}
    \textbf{Output Constraint (JSON):} \\
    Return ONLY a JSON object matching the schema:
    \begin{itemize}
        \setlength\itemsep{0em}
        \item \texttt{doc\_stats}: total sentences and chunk count.
        \item \texttt{chunks}: array of objects, each containing:
        \begin{itemize}
             \item \texttt{id} and \texttt{span} (start/end indices).
             \item \texttt{summary}: string (local summary).
             \item \texttt{contexts}: array of strings (supporting facts).
        \end{itemize}
    \end{itemize}
    
    \vspace{0.5em}
    \textbf{[Begin Document]} \\
    \{\textit{Input Document Content Here}\} \\
    \textbf{[End Document]}
    \end{tcolorbox}

    \vspace{1em}

    \begin{tcolorbox}[title=Decomposition (Deco) Step 2: Merging]
    \small
    \textbf{Instruction:} \\
    You are a merger that produces a single high-quality abstractive summary.
    Merge the given chunk summaries into one single concise summary that covers all key information.
    
    \vspace{0.5em}
    \textbf{Reasoning Guidance:} \\
    Use the provided \textit{chunk summaries} as the primary gist.
    Use the \textit{contexts} to proofread, resolve conflicts, keep terminology consistent, and maintain a clear overall chronology.
    
    \vspace{0.5em}
    \textbf{Formatting Constraint:} \\
    Output ONLY the final summary text (no lists, no JSON, no mention of 'document' or 'chunk').
    
    \vspace{0.5em}
    \textbf{Input Context:} \\
    Below are summaries of different parts:\\
    \{\textit{List of Chunk Summaries (JSON from Step 1)}\}
    
    \vspace{0.5em}
    Below are the supporting contexts for the summaries:\\
    \{\textit{List of Chunk Contexts (JSON from Step 1)}\}
    
    \vspace{1em}
 
    SUMMARY:
    \end{tcolorbox}
    \captionof{figure}{Prompts used for the Decomposition (Deco) method. \textbf{Top:} Step 1 prompt for decomposing the document and generating local summaries with contexts. \textbf{Bottom:} Step 2 prompt for merging the local outputs into a final coherent summary.}\label{fig:prompt_deco}

\newpage
\subsection*{Question-Answer Guided (QAG)}
\begin{tcolorbox}[title=QAG Step 1: Question Generation]
\small
\textbf{Instruction:} \\
You are a question planner for summarization. Read the document and write the exact questions you need to ask to extract its summary facts.

\vspace{0.5em}
\textbf{Task Guidelines:} \\
Propose a focused set of concrete, answerable questions (4--8 adaptively). Aim for coverage first, then depth. Consider the following facets:
\begin{itemize}
    \setlength\itemsep{0em}
    \item topic, key\_pts, entities, timeline
    \item numbers, outcomes, challenges, insights
\end{itemize}

\vspace{0.5em}
\textbf{Output Constraint (JSON):} \\
Return ONLY a JSON object with a \texttt{questions} array (keys: \texttt{id}, \texttt{facet}, \texttt{question}).

\vspace{0.5em}
\textbf{[Begin Document]} \\
\{\textit{Input Document Content Here}\} \\
\textbf{[End Document]}
\end{tcolorbox}

\vspace{1em}

\begin{tcolorbox}[title=QAG Step 2: Answering]
\small
\textbf{Instruction:} \\
You answer the proposed questions using ONLY the given context.
Answer each question precisely and concisely. Keep numbers/dates/entities exact; avoid speculation.

\vspace{0.5em}
\textbf{Output Constraint (JSON):} \\
Return ONLY a JSON object with an \texttt{answers} array. Each item must contain:
\begin{itemize}
    \setlength\itemsep{0em}
    \item \texttt{id} and \texttt{question} (from input).
    \item \texttt{answer}: string.
    \item \texttt{confidence}: integer (1--5).
\end{itemize}

\vspace{0.5em}
\textbf{Input Context:} \\
DOCUMENT: \{\textit{Input Document Content Here}\} \\
QUESTIONS: \{\textit{Question List JSON from Step 1}\}
\end{tcolorbox}

\vspace{1em}

\begin{tcolorbox}[title=QAG Step 3: Summarization]
\small
\textbf{Instruction:} \\
You are a summarizer. Use the Q\&A table as a structured guide to coverage, and the DOCUMENT as the factual ground truth.

\vspace{0.5em}
\textbf{Requirements:} \\
\begin{itemize}
    \setlength\itemsep{0em}
    \item \textbf{Coverage}: reflect high-confidence answers across facets.
    \item \textbf{Faithfulness}: facts must be supported by the DOCUMENT; resolve contradictions using the DOCUMENT.
    \item \textbf{Coherence}: clear flow, consistent terminology.
\end{itemize}

\vspace{0.5em}
\textbf{Input Context:} \\
Q\&A TABLE: \{\textit{Answers JSON from Step 2}\} \\
DOCUMENT: \{\textit{Input Document Content Here}\}

\vspace{1em}
 
    SUMMARY:
\end{tcolorbox}
\captionof{figure}{Prompts used for the Question-Answer Guided (QAG) method. \textbf{Step 1} generates guiding questions based on specific facets. \textbf{Step 2} answers these questions using the source text. \textbf{Step 3} synthesizes the final summary using the Q\&A pairs as a scaffold and the document for factual grounding.}\label{fig:prompt_qag}

\subsection*{Plan-then-Write (Plan)}
\begin{tcolorbox}[title=Plan-then-Write (Plan) Step 1: Planning]
\small
\textbf{Instruction:} \\
You are a summarization planner for a single-document abstractive summary.
Infer the document's domain/genre from content and design a minimal plan accordingly.

\vspace{0.5em}
\textbf{Task Guidelines:} \\
\begin{itemize}
    \setlength\itemsep{0em}
    \item Identify the \textbf{domain} (e.g., news, scientific, narrative).
    \item Clarify the \textbf{goal} and intended \textbf{audience}.
    \item Decide a suitable writing \textbf{style} (e.g., 'concise, neutral').
    \item Provide 3--5 categories of \textbf{salient info} to cover.
    \item Provide an adaptive \textbf{length\_guidance} based on density/complexity.
\end{itemize}

\vspace{0.5em}
\textbf{Output Constraint (JSON):} \\
Return ONLY a valid JSON object with keys: \texttt{domain}, \texttt{goal}, \texttt{audience}, \texttt{style}, \texttt{salient\_info}, \texttt{length\_guidance}.

\vspace{0.5em}
\textbf{[Begin Document]} \\
SOURCE (truncated if long): \\
\{\textit{Input Document Content Here}\}
\end{tcolorbox}

\vspace{1em}

\begin{tcolorbox}[title=Plan-then-Write (Plan) Step 2: Execution]
\small
\textbf{Instruction:} \\
\textbf{Instruction:} \\
You are a skilled analyst tasked with producing a concise and accurate summary from the given text.
The summary must preserve all essential facts, maintain logical flow, and avoid introducing information not present in the text. Follow the provided PLAN strictly. Produce an abstractive summary with adaptive length.

\vspace{0.5em}
\textbf{Requirements (Derived from Plan):} \\
\begin{itemize}
    \setlength\itemsep{0em}
    \item \textbf{Domain context}: \{\textit{domain}\}
    \item \textbf{Style}: \{\textit{style}\}
    \item \textbf{Salient categories}: \{\textit{salient\_info}\}
    \item \textbf{Length guidance}: \{\textit{length\_guidance}\}
\end{itemize}

\vspace{0.5em}
\textbf{Formatting Constraint:} \\
Do NOT output lists, bullets, headings, or metadata. Output the summary text only.
Prefer the minimum length that fully covers the salient information.

\vspace{0.5em}
\textbf{Input Context:} \\
DOCUMENT: \{\textit{Input Document Content Here}\} \\
PLAN: \{\textit{Generated Plan JSON from Step 1}\}

\vspace{1em}
 
    SUMMARY:
\end{tcolorbox}
\captionof{figure}{Prompts used for the Plan-then-Write (Plan) method. \textbf{Step 1} prompts the model to generate a structural plan including domain, style, and salient information. \textbf{Step 2} uses this plan to guide the generation of the final abstractive summary.}\label{fig:prompt_plan}

\newpage
\subsection*{Iterative Refinement (IR)}
\begin{tcolorbox}[title=Iterative Refinement (IR) Step 1: Drafting]
\small
\textbf{Instruction:} \\
You are a skilled analyst tasked with producing a concise and accurate summary from the given text.
The summary must preserve all essential facts, maintain logical flow, and avoid introducing information not present in the text.

\vspace{0.5em}
\textbf{Formatting Constraint:} \\
Then write ONLY the final summary text (no lists, no bullets, no metadata).
Aim to be as concise as possible while fully covering the key points.

\vspace{0.5em}
\textbf{[Begin Document]} \\
\{\textit{Input Document Content Here}\} \\
\textbf{[End Document]}

\vspace{1em}
 
    SUMMARY:
\end{tcolorbox}
\captionof{figure}{Prompts used for the Iterative Refinement (IR) method. \textbf{Step 1} generates an initial draft. \textbf{Step 2} evaluates the draft and provides structured feedback (JSON). \textbf{Step 3} revises the summary based on the feedback. Steps 2 and 3 repeat until the stop condition is met.}\label{fig:prompt_ir}

\vspace{1em}

\begin{tcolorbox}[title=Iterative Refinement (IR) Step 2: Evaluating]
\small
\textbf{Instruction:} \\
You are a summary evaluator. Give a score (1--5) for the summary and provide precise revise suggestions (add, remove, rephrase, shorten, or none).

\vspace{0.5em}
\textbf{Output Constraint (JSON):} \\
Return ONLY a JSON object with:
\begin{itemize}
    \setlength\itemsep{0em}
    \item \texttt{score}: integer 1--5.
    \item \texttt{suggestions}: array of objects (\texttt{type}, \texttt{content}, \texttt{evidence}).
    \item \texttt{stop}: boolean (true if score is 5 and no revision needed).
\end{itemize}

\vspace{0.5em}
\textbf{Input Context:} \\
DOCUMENT: \{\textit{Input Document Content Here}\} \\
CURRENT\_SUMMARY: \{\textit{Summary from Previous Step}\}
\end{tcolorbox}

\vspace{1em}

\begin{tcolorbox}[title=Iterative Refinement (IR) Step 3: Refining]
\small
\textbf{Instruction:} \\
Revise the CURRENT\_SUMMARY according to the EVALUATION suggestions.

\vspace{0.5em}
\textbf{Rules:} \\
\begin{itemize}
    \setlength\itemsep{0em}
    \item Apply actionable suggestions faithfully (add/remove/rephrase/shorten).
    \item If any suggestion introduces hallucination, ignore it.
    \item Output ONLY the revised summary text.
\end{itemize}

\vspace{0.5em}
\textbf{Input Context:} \\
DOCUMENT: \{\textit{Input Document Content Here}\} \\
CURRENT\_SUMMARY: \{\textit{Summary from Previous Step}\} \\
EVALUATION (JSON): \{\textit{Suggestions from Step 2}\}

\vspace{1em}
 
    SUMMARY:
\end{tcolorbox}

\newpage
\subsection*{Extract-to-Abstract (E2A)}
    \begin{tcolorbox}[title=Extract-to-Abstract (E2A) Step 1: Extraction]
    \small
    \textbf{Instruction:} \\
    You are an extractive selector. Your job is to pick the most salient sentences from a document.
    
    \vspace{0.5em}
    \textbf{Selection Policy:} \\
    \begin{itemize}
        \setlength\itemsep{0em}
        \item Split document into sentences ($N$).
        \item Decide selection budget $K$ based on $N$ (e.g., if $N \le 6 \to K=\min(N, 3)$; if $N > 60 \to K=14$).
        \item Select $K$ sentences that maximize salience and minimize redundancy.
        \item Keep each selected sentence \textbf{EXACTLY} as it appears (verbatim).
    \end{itemize}
    
    \vspace{0.5em}
    \textbf{Output Constraint (JSON):} \\
    Return ONLY a valid JSON object with keys:
    \begin{itemize}
        \setlength\itemsep{0em}
        \item \texttt{stats}: object with \texttt{total\_sentences} and \texttt{selected\_budget}.
        \item \texttt{selected}: array of objects (\texttt{index}, \texttt{text}).
    \end{itemize}
    
    \vspace{0.5em}
    \textbf{[Begin Document]} \\
    \{\textit{Input Document Content Here}\} \\
    \textbf{[End Document]}
    \end{tcolorbox}

    \vspace{1em}

    \begin{tcolorbox}[title=Extract-to-Abstract (E2A) Step 2: Abstraction]
    \small
    \textbf{Instruction:} \\
You are a skilled analyst tasked with producing a concise and accurate summary from the given text.
The summary must preserve all essential facts, maintain logical flow, and avoid introducing information not present in the text.
    
    \vspace{0.5em}
    \textbf{Goal:} \\
    \begin{itemize}
        \setlength\itemsep{0em}
        \item Preserve essential facts supported by EVIDENCE (and consistent with DOCUMENT).
        \item Maintain logical flow and coherence.
        \item Avoid copying sentences verbatim (paraphrase).
    \end{itemize}
    
    \vspace{0.5em}
    \textbf{Input Context:} \\
    EVIDENCE: \{\textit{Extracted Sentences JSON from Step 1}\} \\
    DOCUMENT: \{\textit{Input Document Content Here}\}
    
    \vspace{1em}
 
    SUMMARY:
    \end{tcolorbox}
    \captionof{figure}{Prompts used for the Extract-to-Abstract (E2A) method. \textbf{Top:} Step 1 selects salient sentences based on an adaptive budget policy. \textbf{Bottom:} Step 2 generates the abstractive summary conditioned on both the extracted evidence and the full document context.}\label{fig:prompt_e2a}

\fussy

\end{document}